\documentclass[10pt,journal,compsoc]{IEEEtran}

\ifCLASSINFOpdf
\else
\fi
\usepackage{booktabs}
\usepackage{multirow}
\usepackage{times}
\usepackage{xcolor}
\usepackage{soul}
\usepackage[utf8]{inputenc}
\usepackage{epsfig}
\usepackage{graphicx}
\usepackage{amsmath}
\usepackage{amssymb}
\usepackage{url}  %Required
\usepackage{amsmath}
\usepackage{adjustbox}
\usepackage{subfigure}
\usepackage{blindtext}
\usepackage{enumerate}
\usepackage{bm}
\usepackage{dsfont}
\usepackage{array}
\usepackage[numbers]{natbib}
\usepackage[hidelinks]{hyperref}
\usepackage{balance}
\usepackage[ruled,lined,linesnumbered]{algorithm2e}
\usepackage{amsfonts}
\usepackage{setspace}
\usepackage{ragged2e}
\usepackage{lineno}
\usepackage{comment}
\usepackage{color}
\usepackage{tabularx}
\usepackage{makecell}
\usepackage{color, colortbl}
\usepackage{pifont}
\usepackage{hhline}

\renewcommand{\raggedright}{\leftskip=0pt \rightskip=0pt plus 0cm}
\newcolumntype{L}[1]{>{\raggedright\arraybackslash}p{#1}}
\newcolumntype{C}[1]{>{\centering\arraybackslash}p{#1}}
\newcolumntype{R}[1]{>{\raggedleft\arraybackslash}p{#1}}

\definecolor{ForestGreen}{rgb}{0.13, 0.55, 0.13}
\definecolor{Maroon}{rgb}{0.69, 0.19, 0.0}
\definecolor{Gray}{gray}{0.8}

\newcommand{\cmark}{\textcolor{ForestGreen}{\ding{51}}}%
\newcommand{\xmark}{\textcolor{Maroon}{\ding{55}}}%

\newcommand*\rot{\rotatebox{90}}

\begin{document}
%
% paper title
% Titles are generally capitalized except for words such as a, an, and, as,
% at, but, by, for, in, nor, of, on, or, the, to and up, which are usually
% not capitalized unless they are the first or last word of the title.
% Linebreaks \\ can be used within to get better formatting as desired.
% Do not put math or special symbols in the title.
\title{MADAN: Multi-source Adversarial Domain Aggregation Network for Domain Adaptation}

\author{Sicheng~Zhao,\IEEEmembership{~Senior~Member,~IEEE},~Bo~Li,~Xiangyu~Yue,~Pengfei~Xu,~Kurt~Keutzer,\IEEEmembership{~Fellow,~IEEE}
\IEEEcompsocitemizethanks{\IEEEcompsocthanksitem S. Zhao, B. Li, X. Yue, and K. Keutzer are with Department of Electrical Engineering and Computer Sciences, University of California, Berkeley, USA (e-mail: schzhao@gmail.com; drluodian@gmail.com; xyyue@berkeley.edu;  keutzer@berkeley.edu).\protect
\IEEEcompsocthanksitem P. Xu is with Didi Chuxing, Beijing, China (e-mail: xupengfeipf@didiglobal.com).
%\IEEEcompsocitemizethanks{\IEEEcompsocthanksitem S. Zhao, X. Yue, T. Darrell, and K. Keutzer are with Department of Electrical Engineering and Computer Sciences, University of California, Berkeley, USA (e-mail: schzhao@gmail.com; xyyue@berkeley.edu; trevor@eecs.berkeley.edu; keutzer@berkeley.edu).\protect
%\IEEEcompsocthanksitem B. Li and P. Xu are with Didi Chuxing, Beijing, China (e-mail: drluodian@gmail.com; xupengfeipf@didiglobal.com).
}
%\thanks{Copyright (c) 2013 IEEE. Personal use of this material is permitted. However, permission to use this material for any other purposes must be obtained from the IEEE by sending a request to pubs-permissions@ieee.org.}
\thanks{Manuscript received February 18, 2020.}
}

\markboth{Zhao \MakeLowercase{\textit{et al.}}: MADAN}%
{Zhao \MakeLowercase{\textit{et al.}}: MADAN}

\IEEEtitleabstractindextext{%
\begin{abstract}
Domain adaptation aims to learn a transferable model to bridge the domain shift between one labeled source domain and another sparsely labeled or unlabeled target domain. Since the labeled data may be collected from multiple sources, multi-source domain adaptation (MDA) has attracted increasing attention. Recent MDA methods do not consider the pixel-level alignment between sources and target or the misalignment across different sources. In this paper, we propose a novel MDA framework to address these challenges. Specifically, we design an end-to-end Multi-source Adversarial Domain Aggregation Network (MADAN). First, an adapted domain is generated for each source with \textit{dynamic semantic consistency} while aligning towards the target at the pixel-level cycle-consistently. Second, \textit{sub-domain aggregation discriminator} and \textit{cross-domain cycle discriminator} are proposed to make different adapted domains more closely aggregated. Finally, feature-level alignment is performed between the aggregated domain and the target domain while training the task network. For the segmentation adaptation, we further enforce \textit{category-level alignment} and incorporate \textit{context-aware generation}, which constitutes MADAN+. We conduct extensive MDA experiments on digit recognition, object classification, and simulation-to-real semantic segmentation. The results demonstrate that the proposed MADAN and MANDA+ models outperform state-of-the-art approaches by a large margin.
\end{abstract}
}

% make the title area
\maketitle

\IEEEdisplaynontitleabstractindextext
% \IEEEdisplaynontitleabstractindextext has no effect when using
% compsoc or transmag under a non-conference mode.

\IEEEpeerreviewmaketitle

\IEEEraisesectionheading{\section{Introduction}\label{sec:Introduction}}
%\hfill mds
%\hfill August 26, 2015

\begin{figure}[!t]
\begin{center}
\centering \includegraphics[width=1.0\linewidth]{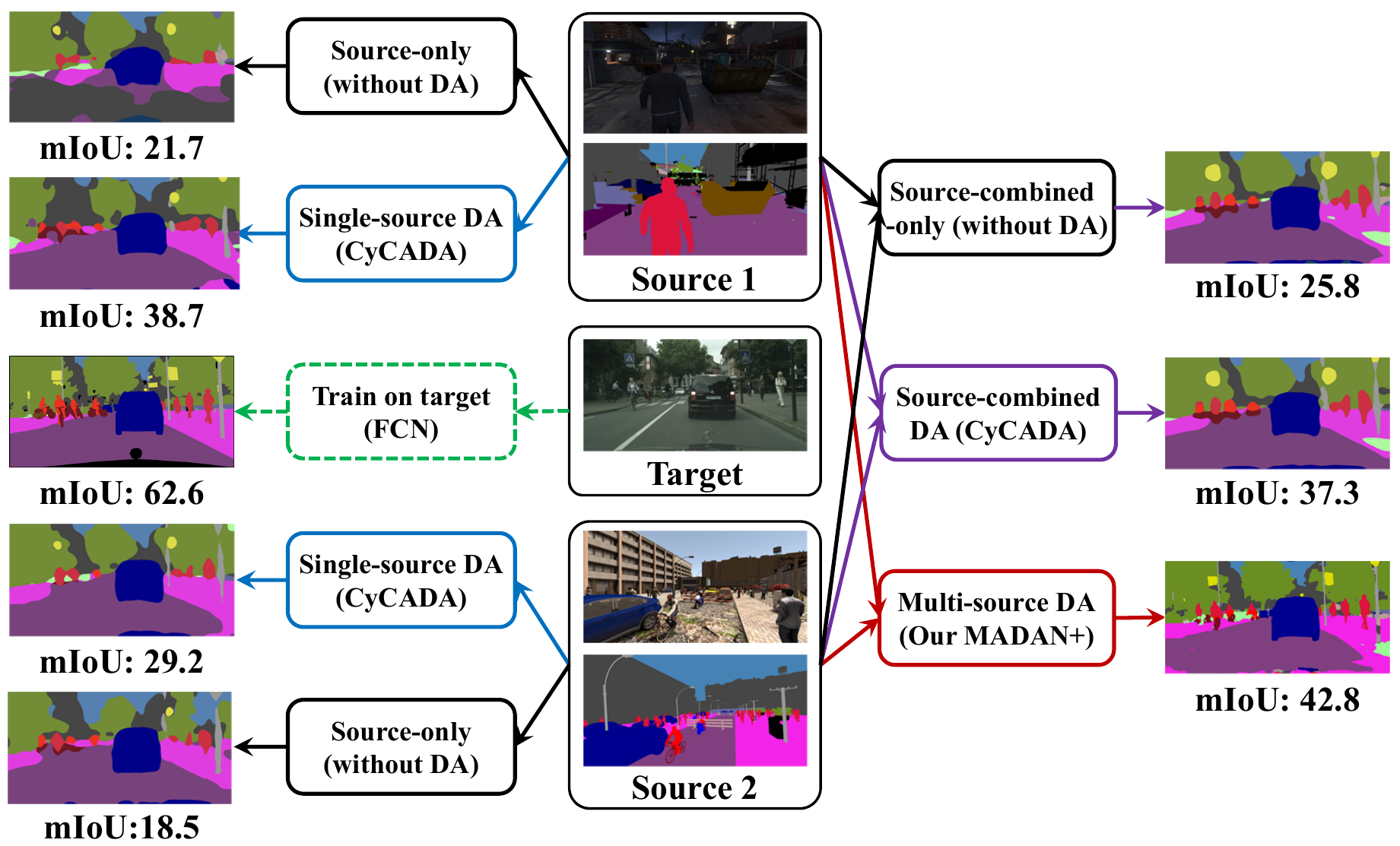}
\caption{An example of \emph{domain shift}. Source 1: GTA, Source 2: SYNTHIA, Target: Cityscapes. Left three columns: single-source DA. The overall mIoU result of the FCN semantic segmentation model\protect~\cite{long2015fully} drops from 62.6\% (trained on the target Cityscapes, unavailable in UDA and simply used for comparison here) to 21.7\% and 18.5\% (trained only on the source GTA and SYNTHIA). CyCADA achieves 38.7\% and 29.2\%, demonstrating that DA can bridge the domain gap.
Right three columns: multi-source DA. Simply combining multiple sources and performing single-source DA (37.3\%) does not outperform the best single-source DA (38.7\%). We propose Multi-source Adversarial Domain Aggregation Network (MADAN), a novel adversarial model, to perform multi-source DA. Our method achieves significant performance improvements over source-combined DA,  source-combined only, and single-source DA.}
\label{fig:DomainShift}
\end{center}
\end{figure}

\IEEEPARstart{T}{ogether} with increased computation capacity and deep complex models, large-scale labeled data attributes to the significant success of deep learning algorithms as one key element. Consequently, promising performance has been obtained via deep neural networks in various computer vision tasks, such as image classification~\cite{krizhevsky2012imagenet,simonyan2014very,he2016deep,huang2017densely}, object detection~\cite{girshick2015fast,ren2015faster,redmon2016you}, and semantic segmentation~\cite{long2015fully,badrinarayanan2017segnet,chen2017deeplab}. However, in many real-world applications, there are only limited or even no labeled training data, as labeling is expensive, time-consuming, and difficult. For example, only the labels provided by experts are reliable in fine-grained recognition~\cite{gebru2017fine}; labeling each Cityscapes image takes about 90 minutes in semantic segmentation~\cite{cordts2016cityscapes}; point-wise 3D LiDAR point clouds are difficult to label in autonomous driving~\cite{wu2019squeezesegv2,yue2018lidar}. One direct way is to transfer the learned knowledge from one labeled source domain to another different but related target domain. However, because of the presence of \emph{domain shift} or \emph{dataset bias}~\cite{torralba2011unbiased}, \textit{i.e.} the joint probability distributions of observed data and labels are different in the two domains, direct transfer may not perform well, as shown in Figure~\ref{fig:DomainShift}. This observation motivates the research on domain adaptation (DA)~\cite{bousmalis2016domain,tzeng2017adversarial}.

%The left three columns and right three column correspond to the domain shift for single-source DA and multi-source DA, respectively.

Without requiring any labeled data from the target domain, unsupervised domain adaptation (UDA) is the most widely studied pipeline. Both theoretical analysis~\cite{ben2010theory,gopalan2014unsupervised,louizos2015variational,tzeng2017adversarial} and algorithm design~\cite{pan2010survey,glorot2011domain,jhuo2012robust,becker2013non,ghifary2015domain,long2015learning,hoffman2018cycada,zhao2019cycleemotiongan} for UDA have been proposed recently. Conventional UDA methods mainly focus on the single-source scenario based on the assumption that the labeled source data is sampled from the same distribution. However, in practice, the labeled data may be collected from multiple sources with different distributions~\cite{sun2015survey,zhao2020multi}. Simply combining different sources into one source and directly employing single-source UDA may lead to suboptimal solutions, since the data from different sources may interfere with each other during the learning process~\cite{riemer2019learning}, as shown in Figure~\ref{fig:DomainShift}. Therefore, effective multi-source domain adaptation (MDA) algorithms are required.

Early efforts on MDA mainly used shallow models~\cite{sun2015survey},
either learning a latent feature space for different domains~\cite{duan2009domain,sun2011two,duan2012exploiting,chattopadhyay2012multisource,duan2012domain} or combining pre-learned source classifiers~\cite{yang2007cross,schweikert2009empirical,xu2012multi,sun2013bayesian}. Recently, some deep MDA methods that only focus on image classification have been proposed by learning a common feature space and aligning each source and target pair~\cite{xu2018deep,zhao2018adversarial,peng2019moment,zhao2020multi}. There are some limitations of these methods. (1) They mainly focus on feature-level alignment, which only aligns high-level information. This might be sufficient for coarse-grained classification tasks, but it is obviously insufficient for fine-grained semantic segmentation, which performs pixel-wise prediction. Further, they have low interpretability, which cannot well explain why these methods work. (2) They only align each source and target pair. Although different sources are matched towards the target, there may exist significant mis-alignment across different sources. (3) They only focus on image classification where one label is assigned to each image. Directly extending them from classification to segmentation, which assigns a semantic label (\textit{e.g.} car, cyclist, pedestrian, road) to each pixel in an image, may not perform well. This is because segmentation is a structured prediction task, \textit{i.e.} it has to resolve the predictions in an exponentially large label space and thus the decision function is more involved than classification~\cite{zhang2017curriculum,tsai2018learning}.

To address the above challenges, in this paper we propose a novel MDA framework, termed Multi-source Adversarial Domain Aggregation Network (MADAN), which consists of Dynamic Adversarial Image Generation, Adversarial Domain Aggregation, and Feature-aligned task learning. First, for each source, we generate an adapted domain using a Generative Adversarial Network (GAN)~\cite{goodfellow2014generative} with cycle-consistency constraint~\cite{zhu2017unpaired}, which enforces pixel-level alignment between source images and target images. To preserve the semantics before and after image translation, we propose a novel semantic consistency loss by minimizing the Kullback–Leibler (KL) divergence between the source predictions of a pretrained task model (\textit{e.g.} classification and segmentation) and the adapted predictions of a \textit{dynamic task model}. Second, instead of training a classifier for each source domain~\cite{xu2018deep,peng2019moment,zhao2020multi}, we propose \textit{sub-domain aggregation discriminator} to directly make different adapted domains indistinguishable, and \textit{cross-domain cycle discriminator} to discriminate between the images from each source and the images transferred from other sources. In this way, different adapted domains can be better aggregated into a more unified domain.
Finally, the task model is trained on the aggregated domain, while enforcing feature-level alignment between the aggregated domain and the target domain.

In summary, our contributions are three-fold:
\begin{itemize}
    \item We design a novel framework termed MADAN to do multi-source domain adaptation. (i) Sub-domain aggregation discriminator and cross-domain cycle discriminator are proposed to better align different adapted domains. (ii) Besides feature-level alignment, pixel-level alignment is further considered by generating an adapted domain for each source cycle-consistently with a novel dynamic semantic consistency loss.
    \item We propose to perform domain adaptation for semantic segmentation from multiple sources. To our best knowledge, this is the first work on multi-source structured domain adaptation. For segmentation, MADAN is enhanced to MADAN+ with category-level alignment and context-aware generation.
    \item We conduct extensive experiments on several MDA bechmark datasets for digit recognition, object classification, and simulation-to-real semantic segmentation, and the results demonstrate the effectiveness of the proposed MADAN and MADAN+ models.
\end{itemize}

One preliminary version on MADAN was previously introduced in our NeurIPS conference paper~\cite{zhao2019multi}. As compared to the conference version, this journal paper has the following three aspects of enhancements. First, we perform a more comprehensive review to compare the proposed method with existing methods. Second, we conduct MDA experiments on digit recognition and object classification, which also achieves state-of-the-art performances, and enrich the analysis of the results. Third, we extend the original MADAN to MADAN+ with category-level alignment and context-aware generation for semantic segmentation, conduct more comparative experiments, and achieve better performances.

The rest of this paper is organized as follows. Section~\ref{sec:RelatedWork} reviews related work on single-source UDA and MDA. Section~\ref{sec:ProblemSetup} gives the definition of the MDA problem. Section~\ref{sec:MADAN} describes the proposed MADAN and extended MADAN+ models in detail. Experimental settings, results, and analysis are presented in Section~\ref{sec:Experiments}. We conclude this paper in Section~\ref{sec:Conclusion}.

\begin{table*}[!t]
\centering\footnotesize%\small%\scriptsize%
\caption{Comparison of the proposed MADAN model with several state-of-the-art domain adaptation methods. The full names of each property from the third to the last columns are pixel-level alignment, context-aware generation, feature-level alignment, category-level alignment, semantic consistency, cycle consistency, multiple sources, domain aggregation, one task network, fine-grained prediction, and end-to-end training, respectively.}
\begin{tabular}
{c c c c c c c c c c c c c c}
\toprule
%& pixel-level & feature-level & semantic & cycle & multiple & domain & number of & fine-grained \\
%& alignment & alignment & consistency & consistency & sources & aggregation & task networks & prediction \\
DA setting & method & pixel & con & feat & cat & sem & cycle & multi & aggr & one & fine & end\\
\hline
& ADDA~\cite{tzeng2017adversarial} & \xmark  & \xmark  & \cmark & \xmark  & -- &  -- & \xmark  & -- & \cmark & \cmark &  \xmark \\
& CycleGAN~\cite{zhu2017unpaired} & \cmark  & \xmark  & \xmark & \xmark  & \xmark &  \cmark & \xmark  & -- & \cmark & \xmark &  \xmark  \\
& PixelDA~\cite{bousmalis2017unsupervised} & \cmark & \xmark  & \xmark & \xmark  & \xmark &  \xmark & \xmark  & -- & \cmark & \cmark  & \cmark \\
& NMD~\cite{chen2017no}  & \xmark & \xmark  & \cmark & \cmark  & - &  - & \xmark  & -- & \cmark & \cmark  & \cmark \\
single-source& SBADA~\cite{russo2018source} & \cmark  & \xmark  & \xmark & \xmark  & \cmark &  \cmark & \xmark  & -- & \cmark & \xmark  & \cmark \\
& GTA-GAN~\cite{sankaranarayanan2018generate} & \cmark  & \xmark  & \cmark & \xmark  & \xmark &  \xmark & \xmark  & -- & \cmark & \xmark  & \cmark \\
& DupGAN~\cite{hu2018duplex} & \cmark  & \xmark  & \cmark & \xmark  & \cmark &  \xmark & \xmark  & -- & \cmark & \xmark &  \cmark \\
& CyCADA~\cite{hoffman2018cycada} & \cmark  & \xmark  & \cmark & \xmark  & \cmark &  \cmark & \xmark  & -- & \cmark & \cmark &  \cmark  \\
\hline
& DCTN~\cite{xu2018deep} & \xmark  & \xmark  & \cmark & \xmark  & -- &  -- & \cmark  & \xmark & \xmark & \xmark &  \cmark \\
& MDAN~\cite{zhao2018adversarial} & \xmark & \xmark   & \cmark & \xmark  &  -- &  -- & \cmark  & \xmark & \cmark & \xmark &  \cmark \\
multi-source& MMN~\cite{peng2019moment} & \xmark & \xmark   & \cmark & \xmark  &  -- &  -- & \cmark  & \xmark & \xmark & \xmark  & \cmark \\
& MDDA~\cite{zhao2020multi} & \xmark & \xmark   & \cmark & \xmark  &  -- & -- & \cmark  & \xmark & \xmark & \xmark  & \xmark \\
& MADAN (ours) & \cmark  & \xmark  & \cmark & \xmark  &  \cmark &  \cmark & \cmark  & \cmark & \cmark & \cmark  & \cmark \\
& MADAN+ (ours) & \cmark  & \cmark  & \cmark & \cmark  &  \cmark &  \cmark & \cmark  & \cmark & \cmark & \cmark  & \cmark \\
\bottomrule
\end{tabular}
\label{tab:propertyComparison}
\end{table*}

\section{Related Work}
\label{sec:RelatedWork}

In this section, we introduce related work on single-source unsupervised domain adaptation and multi-source domain adaptation, and compare the proposed MADAN with these methods.

\subsection{Single-source UDA}

While the early single-source UDA (SUDA) methods are mainly non-deep ones~\cite{patel2015visual}, either re-weighting samples or transforming intermediate subspaces, the emphasis of recent SUDA methods has shifted to deep learning architectures in an end-to-end fashion. Typically, a conjoined architecture with two streams is employed in deep SUDA~\cite{zhuo2017deep}. One stream is used to represent the task model for the source domain, and the other is used to align the target and source domains. Correspondingly, a traditional task loss based on the labeled source data and another alignment loss to tackle the domain shift problem are jointly optimized during the training of deep SUDA. Typically, the task loss is the same among different methods, while the difference is focused on the alignment loss, such as discrepancy loss, adversarial loss, reconstruction loss, \emph{etc}.

%correlation alignment (CORAL)~\cite{sun2016return,sun2017correlation,zhuo2017deep}, Geodesic distance~\cite{wu2019squeezesegv2},

Discrepancy-based methods explicitly measure the discrepancy between the target domain and the source domain, such as the multiple kernel variant of maximum mean discrepancies~\cite{long2015learning}, correlation alignment (CORAL)~\cite{sun2016return,zhuo2017deep}, geodesic distance~\cite{wu2019squeezesegv2}, and contrastive domain discrepancy~\cite{kang2019contrastive}. Adversarial generative methods combine the domain discriminative model with a generative component to generate fake source or target data generally based on GAN~\cite{goodfellow2014generative,bousmalis2017unsupervised} and its variants, such as CoGAN~\cite{liu2016coupled}, SimGAN~\cite{shrivastava2017learning}, CycleGAN~\cite{zhu2017unpaired,zhao2019cycleemotiongan,yue2019domain}, and CyCADA~\cite{hoffman2018cycada}. Adversarial discriminative methods usually employ an adversarial objective with respect to a domain discriminator to encourage domain confusion~\cite{ganin2016domain,tzeng2017adversarial,chen2017no,shen2017wasserstein,tsai2018learning,huang2018domain}.
Reconstruction based methods try to reconstruct the target input from the features extracted using the source task model by minimizing the reconstruction loss~\cite{ghifary2015domain,ghifary2016deep}. While the adversarial generative methods consider the pixel-level alignment, the others mainly employ feature-level alignment. Although these methods make remarkable progress to SUDA, they suffer from large performance decay when directly applied to the MDA problem.

\subsection{Multi-source Domain Adaptation}

Multi-source domain adaptation (MDA) considers a more practical scenario, where the training data are collected from multiple sources~\cite{sun2015survey,zhao2019multi}. Some theoretical analysis~\cite{ben2010theory,hoffman2018algorithms} is developed to support existing MDA algorithms. The early MDA methods mainly focus on shallow models, including two categories~\cite{sun2015survey}: feature representation approaches~\cite{duan2009domain,sun2011two,duan2012exploiting,chattopadhyay2012multisource,duan2012domain} and combination of pre-learned classifiers~\cite{yang2007cross,schweikert2009empirical,xu2012multi,sun2013bayesian}. Some recent shallow MDA methods mainly aim to deal with special cases, such as incomplete MDA~\cite{ding2018incomplete} and target shift~\cite{redko2019optimal}.

Recently, some representative deep learning based MDA methods are proposed, such as multisource domain adversarial network (MDAN)~\cite{zhao2018adversarial}, deep cocktail network (DCTN)~\cite{xu2018deep},  moment matching network (MMN)~\cite{peng2019moment}, and multi-source distilling domain adaptatioin (MDDA)~\cite{zhao2020multi}. All these MDA methods only consider the feature-level alignment for image classification tasks. The former three methods employ a shared feature extractor to symmetrically map the multiple sources and target into the same space. %, without considering the conditional shift problem.
For each source-target pair in MDAN and DCTN, a discriminator is trained to distinguish the source and target features. MDAN directly concatenates all extracted source features and labels into one domain and train a single task model, while a task model is trained for each source domain in DCTN, which combines the predictions of different models for a target image using perplexity scores as weights. MMN transfers the learned knowledge from multiple sources to the target by dynamically aligning moments of their feature distributions. The final prediction of a target image is averaged uniformly based on the classifiers from different source domains. MDDA first pre-trains a feature extractor for each source and match the target feature to each source feature space asymmetrically. After distilling the pre-trained classifiers with selected representative samples in each source, the predictions of the matched target features using corresponding source classifiers are combined based on the weights obtained from the Wasserstein distance. Differently, we also consider the pixel-level alignment. Based on the aggregated intermediate domain obtained by sub-domain aggregation discriminator and cross-domain cycle discriminator, only one task model needs to be trained. Besides the image classification tasks, we also perform semantic segmentation task, which is the first work on MDA for segmentation. Table~\ref{tab:propertyComparison} compares MADAN with several state-of-the-art DA methods.

\begin{figure*}[!t]
\centering
\includegraphics[width=0.98\linewidth]{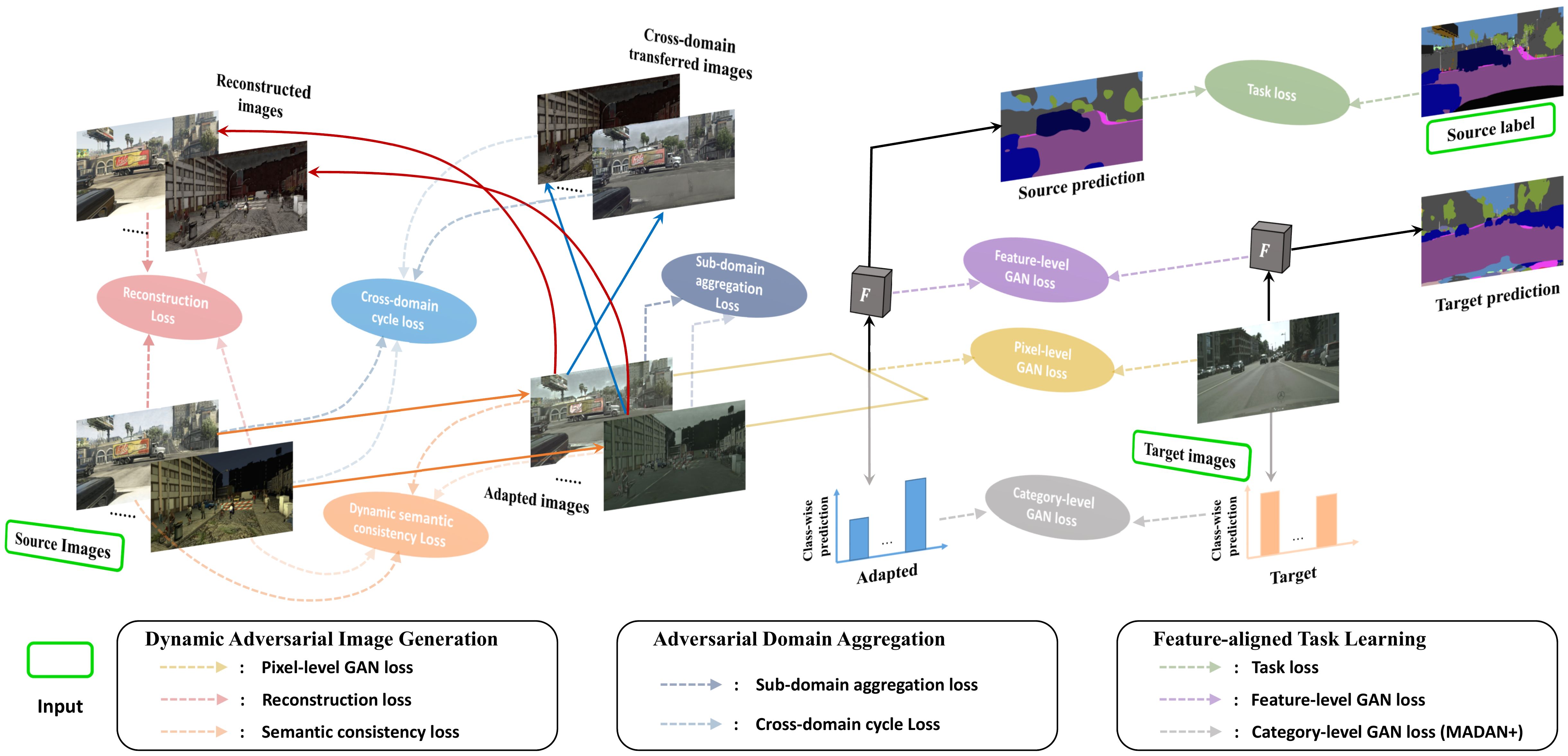}
\caption{The framework of the proposed Multi-source Adversarial Domain Aggregation Network (MADAN). The colored solid arrows represent generators, while the black and grey solid arrows indicate the task network $F$. The dashed arrows correspond to different losses.}
\label{fig:Framework_MADAN}
\end{figure*}

\section{Problem Setup}
\label{sec:ProblemSetup}

We consider the unsupervised MDA scenario with multiple labeled source domains $S_1,S_2,\cdots,S_M$, where $M$ is number of sources, and one unlabeled target domain $T$. In the $i$th source domain $S_i$, suppose $X_i=\{\mathbf{x}_i^j\}_{j=1}^{N_i}$ and $Y_i=\{\mathbf{y}_i^j\}_{j=1}^{N_i}$ are the observed data and corresponding labels drawn from the source distribution $p_i(\mathbf{x}, \mathbf{y})$, where $N_i$ is the number of samples in $S_i$. For different tasks, the format of labels $\mathbf{y}_i^j$ varies. For example, in classification, each image has a unique $\mathbf{y}_i^j$; in segmentation, $\mathbf{y}_i^j$ is pixel-wise. In the target domain $T$, let $X_T=\{\mathbf{x}_T^j\}_{j=1}^{N_T}$ denote the target data drawn from the target distribution $p_T(\mathbf{x},\mathbf{y})$ without label observation, where $N_T$ is the number of target samples.
Unless otherwise specified, we have two assumptions: (1) homogeneity, \textit{i.e.} $\mathbf{x}_i^j\in \mathbb{R}^{d}, \mathbf{x}_T^j\in \mathbb{R}^{d}$, indicating that the data from different domains are observed in the same image space but with different distributions; (2) closed set, \textit{i.e.} $\mathbf{y}_i^j\in \mathcal{Y}, \mathbf{y}_T^j\in \mathcal{Y}$, where $\mathcal{Y}$ is the label set, which means that all the domains share the same space of classes. Based on covariate shift and concept drift~\cite{patel2015visual}, we aim to learn an adaptation model that can correctly predict the labels of a sample from the target domain trained on $\{(X_i,Y_i)\}_{i=1}^{M}$ and $\{X_T\}$. How to extend the unsupervised, homogeneous, and closed set MDA method to other settings, such as heterogeneous DA, open set DA, and category-shift DA remains our future work.

\section{Multi-source Adversarial Domain Aggregation Network}
\label{sec:MADAN}

In this section, we introduce the proposed Multi-source Adversarial Domain Aggregation Network (MADAN) for image classification and semantic segmentation adaptation in detail. The framework is illustrated in Figure~\ref{fig:Framework_MADAN}, which consists of three components: Dynamic Adversarial Image Generation (DAIG), Adversarial Domain Aggregation (ADA), and Feature-aligned Task Learning (FTL). DAIG aims to generate adapted images from source domains to the target domain from the perspective of visual appearance while preserving the semantic information dynamically. In order to reduce the distances among the adapted domains and thus generate a more aggregated unified domain, ADA is proposed, including Cross-domain Cycle Discriminator (CCD) and Sub-domain Aggregation Discriminator (SAD). Finally, FTL learns the domain-invariant representations at the feature-level in an adversarial manner.

\subsection{Dynamic Adversarial Image Generation}
\label{ssec:DAIG}

The goal of DAIG is to make images from different source domains visually similar to the target images, as if they are drawn from the same target domain distribution. To this end, for each source domain $S_i$, we introduce a generator $G_{S_i\rightarrow T}$ mapping to the target $T$ in order to generate adapted images that fool $D_T$, which is a pixel-level adversarial discriminator. $D_T$ is trained simultaneously with each $G_{S_i\rightarrow T}$ to classify real target images $X_T$ from adapted images $G_{S_i\rightarrow T}(X_i)$. The corresponding GAN loss function is:
% \begin{equation}
% \mathcal{L}_{GAN}^{S_i\rightarrow T}(G_{S_{i}\rightarrow T},D_T, X_i, X_T)=\mathbb{E}_{\mathbf{x}_{i}\sim X_i}\log D_T(G_{S_{i}\rightarrow T}(\mathbf{x}_{i}))+\mathbb{E}_{\mathbf{x}_T\sim X_T}\log [1-D_T(\mathbf{x}_T)].\\
% \label{equ:ganSourceSiT}
% \end{equation}
\begin{equation}\small
\begin{aligned}
&\mathcal{L}_{GAN}^{S_i\rightarrow T}(G_{S_{i}\rightarrow T},D_T, X_i, X_T)=\\
&\mathbb{E}_{\mathbf{x}_{i}\sim X_i}\log D_T(G_{S_{i}\rightarrow T}(\mathbf{x}_{i}))+\mathbb{E}_{\mathbf{x}_T\sim X_T}\log [1-D_T(\mathbf{x}_T)].\\
\end{aligned}
\label{equ:ganSourceSiT}
\end{equation}

Since the mapping $G_{S_i\rightarrow T}$ is highly under-constrained~\cite{goodfellow2014generative}, we employ an inverse mapping $G_{T\rightarrow S_i}$ as well as a cycle-consistency loss~\cite{zhu2017unpaired} to enforce $G_{T\rightarrow S_i}(G_{S_i\rightarrow T}(\mathbf{x}_i)) \approx \mathbf{x}$ and vice versa. Similarly, we introduce $D_i$ to classify $X_i$ from $G_{T\rightarrow S_i}(X_T)$, with the following GAN loss:
% \begin{equation}
% \mathcal{L}_{GAN}^{T\rightarrow S_i}(G_{T\rightarrow S_{i}},D_{i}, X_T, X_{i})=\mathbb{E}_{\mathbf{x}_{i}\sim X_{i}}\log [1-D_{i}(\mathbf{x}_{i})]+\mathbb{E}_{\mathbf{x}_t\sim X_T}\log D_{i}(G_{T\rightarrow S_{i}}(\mathbf{x}_t)).\\
% \label{equ:ganSourceTSi}
% \end{equation}
\begin{equation}\small
\begin{aligned}
&\mathcal{L}_{GAN}^{T\rightarrow S_i}(G_{T\rightarrow S_{i}},D_{i}, X_T, X_{i})=\\
&\mathbb{E}_{\mathbf{x}_{i}\sim X_{i}}\log [1-D_{i}(\mathbf{x}_{i})]+\mathbb{E}_{\mathbf{x}_T\sim X_T}\log D_{i}(G_{T\rightarrow S_{i}}(\mathbf{x}_T)).\\
\end{aligned}
\label{equ:ganSourceTSi}
\end{equation}
The cycle-consistency loss~\cite{zhu2017unpaired} ensures that the learned mappings $G_{S_i\rightarrow T}$ and $G_{T\rightarrow S_i}$ are cycle-consistent, thereby preventing them from contradicting each other, is defined as:
\begin{equation}\small
\begin{aligned}
&\mathcal{L}_{cyc}^{S_i\leftrightarrow T}(G_{S_i\rightarrow T},G_{T\rightarrow S_i}, X_{i}, X_T)=\\
&\mathbb{E}_{\mathbf{x}_{i}\sim X_{i}}\parallel G_{T\rightarrow S_{i}}(G_{S_{i}\rightarrow T}(\mathbf{x}_{i}))-\mathbf{x}_{i}\parallel_1+\\
&\mathbb{E}_{\mathbf{x}_T\sim X_T}\parallel G_{S_{i}\rightarrow T}(G_{T\rightarrow S_{i}}(\mathbf{x}_T))-\mathbf{x}_T\parallel_1.
\end{aligned}
\label{equ:cycSourceSiT}
\end{equation}

The adapted images are expected to contain the same semantic information as original source images, but the semantic consistency is only partially constrained by the cycle consistency loss. The semantic consistency loss in CyCADA~\cite{hoffman2018cycada} was proposed to better preserve semantic information. $\mathbf{x}_i$ and $G_{S_i \rightarrow T}(\mathbf{x}_i)$ are both fed into a task model $F_{i}$ pretrained on $(X_i, Y_i)$. However, since $\mathbf{x}_i$ and $G_{S_i \rightarrow T}(\mathbf{x}_i)$ are from different domains, employing the same task model, \textit{i.e.} $F_{i}$, to obtain the predicted results and then computing the semantic consistency loss may be detrimental to image generation. Ideally, the adapted images $G_{S_i \rightarrow T}(\mathbf{x}_i)$ should be fed into a network  $F_T$ trained on the target domain, which is infeasible since target domain labels are not available in UDA. Instead of employing $F_{i}$ on $G_{S_i \rightarrow T}(\mathbf{x}_i)$, we propose to dynamically update the network $F_A$, which takes $G_{S_i \rightarrow T}(\mathbf{x}_i)$ as input, so that its optimal input domain (the domain that the network performs best on) gradually changes from that of $F_i$ to $F_T$. We employ the task model $F$ trained on the adapted domain as $F_A$, \textit{i.e.} $F_A = F$, which has two advantages: (1) $G_{S_i \rightarrow T}(\mathbf{x}_i)$ becomes the optimal input domain of $F_A$, and as $F$ is trained to have better performance on the target domain, the semantic loss after $F_A$ would promote $G_{S_i \rightarrow T}$ to generate images that are closer to target domain at the pixel-level; (2) since $F_A$ and $F$ can share the parameters, no additional training or memory space is introduced, which is quite efficient.
The proposed dynamic semantic consistency (DSC) loss is:
% \begin{equation}
% \mathcal{L}_{sem}^{S_i}(G_{S_{i}\rightarrow T},X_i,F_{i}, F_{A})=\mathbb{E}_{\mathbf{x}_{i}\sim X_i}KL(F_{A}(G_{S_{i}\rightarrow T}(\mathbf{x}_{i}))||F_{i}(\mathbf{x}_{i})),\\
% \label{equ:semConsisSi}
% \end{equation}
\begin{equation}\small
\begin{aligned}
&\mathcal{L}_{DSC}^{S_i}(G_{S_{i}\rightarrow T},X_i,F_{i}, F_{A})=\\
&\mathbb{E}_{\mathbf{x}_{i}\sim X_i}KL(F_{A}(G_{S_{i}\rightarrow T}(\mathbf{x}_{i}))||F_{i}(\mathbf{x}_{i})),\\
\end{aligned}
\label{equ:semConsisSi}
\end{equation}
where $KL(\cdot||\cdot)$ is the KL divergence between two distributions.
%where $KL(p||q)$ is the KL divergence between two distributions $p$ and $q$.

\subsection{Adversarial Domain Aggregation}
We can train different task models for each adapted domain and combine different predictions with specific weights for target images~\cite{xu2018deep,peng2019moment}, or we can simply combine all adapted domains together and train one model~\cite{zhao2018adversarial}. In the first strategy, it is challenging to determine how to select the weights for different adapted domains. Moreover, each target image needs to be fed into all task models at reference time, and this is rather inefficient. For the second strategy, since the alignment space is high-dimensional, although the adapted domains are relatively aligned with the target, they may be significantly misaligned with each other. In order to mitigate this issue, we propose adversarial domain aggregation to make different adapted domains more closely aggregated with two kinds of discriminators. One is the sub-domain aggregation discriminator (SAD), which is designed to directly make the different adapted domains indistinguishable. For $S_i$, a discriminator $D_A^{i}$ is introduced with the following loss function:
\begin{equation}\small
\begin{aligned}
&\mathcal{L}_{SAD}^{S_i}(G_{S_{1}\rightarrow T}, \dots G_{S_{i}\rightarrow T}, \dots,  G_{S_{M}\rightarrow T}, D_A^i)=\\
&\mathbb{E}_{\mathbf{x}_i\sim X_i}\log D_A^i(G_{S_{i}\rightarrow T}(\mathbf{x}_i))+\\
&\frac{1}{M-1}\sum\nolimits_{j\neq i}\mathbb{E}_{\mathbf{x}_j\sim {X_j}} \log[1-D_A^i(G_{S_{j}\rightarrow  T}(\mathbf{x}_{j}))].\\
\end{aligned}
\label{equ:ganD3}
\end{equation}
The other is the cross-domain cycle discriminator (CCD). For each source domain $S_i$, we transfer the images from the adapted domains $G_{S_j\rightarrow T}(X_j),\ j=1,\cdots,M,j\neq i$ back to $S_i$ using $G_{T\rightarrow S_i}$ and employ the discriminator $D_i$ to classify $X_i$ from $G_{T\rightarrow S_i}(G_{S_j\rightarrow T}(X_j))$, which corresponds to the following loss function:
\begin{equation}\small
\begin{aligned}
&\mathcal{L}_{CCD}^{S_i}(G_{T\rightarrow S_1}, \dots G_{T\rightarrow S_{i-1}}, G_{T\rightarrow S_{i+1}}, \dots, G_{T\rightarrow S_{M}}, G_{S_i\rightarrow T},D_i)\\
&=\mathbb{E}_{\mathbf{x}_{i}\sim X_{i}}\log D_i(\mathbf{x}_{i})+\\
&\frac{1}{M-1}\sum\nolimits_{j\neq i}\mathbb{E}_{\mathbf{x}_{j}\sim X_{j}}\log[1-D_i(G_{T\rightarrow S_{i}}((G_{S_{j}\rightarrow T}(\mathbf{x}_{j})))].\\
\end{aligned}
\label{equ:CCDloss}
\end{equation}
Please note that using a more sophisticated combination of different discriminators' losses to better aggregate the domains with larger distances might improve the performance. We leave this as future work and would explore this direction by dynamic weighting of the loss terms and incorporating some prior domain knowledge of the sources.

\subsection{Feature-aligned Task Learning} %Semantic Learning}
\label{ssec:FLA}

After adversarial domain aggregation, the adapted images of different domains $X_{i}'(i=1,\cdots,M)$ are more closely aggregated and aligned. Meanwhile, the semantic consistency loss in dynamic adversarial image generation ensures that the semantic information, \textit{i.e.} the labels, is preserved before and after image translation. Suppose the images of the unified aggregated domain are $X'=\bigcup\limits_{i=1}^{M}X_{i}'$ and corresponding labels are
$Y=\bigcup\limits_{i=1}^{M}Y_i$. We can then train a task learning model $F$ based on $X'$ and $Y$. For classification and segmentation, $F$ aims to respectively minimize the following cross-entropy loss $\mathcal{L}_{task}(F,X',Y)$:
% \begin{equation}
% \mathcal{L}_{task}(F,X',Y)=-\mathbb{E}_{(\mathbf{x}',\mathbf{y})\sim (X^{'},Y)}\sum\nolimits_{l=1}^{L}\sum\nolimits_{h=1}^{H}\sum\nolimits_{w=1}^{W}\mathds{1}_{[l=\mathbf{y}_{h,w}]}\log(\sigma(F_{l,h,w}(\mathbf{x}'))),
% \label{equ:f_loss}
% \end{equation}
\begin{equation}\small
\mathcal{L}_{cla}(F,X',Y)=-\mathbb{E}_{(\mathbf{x}',y)\sim (X^{'},Y)}\sum\nolimits_{l=1}^{L}\mathds{1}_{[l=y]}\log(\sigma(F(\mathbf{x}'))),\\
\label{equ:f_cla_loss}
\end{equation}
\begin{equation}\small
\begin{aligned}
&\mathcal{L}_{seg}(F,X',Y)=-\mathbb{E}_{(\mathbf{x}',\mathbf{y})\sim (X^{'},Y)}\sum\nolimits_{l=1}^{L}\sum\nolimits_{h=1}^{H}\sum\nolimits_{w=1}^{W}\\
&\mathds{1}_{[l=\mathbf{y}_{h,w}]}\log(\sigma(F_{l,h,w}(\mathbf{x}'))),\\
\end{aligned}
\label{equ:f_seg_loss}
\end{equation}
where $L$ is the number of classes, $H,W$ are the height and width of the adapted images, $\sigma$ is the softmax function,  $\mathds{1}$ is an indicator function, and $F_{l,h,w}(\mathbf{x}')$ is the value of $F(\mathbf{x}')$ at index $(l,h,w)$.

Further, we impose feature-level alignment between $X'$ and $X_T$, which can improve the task performance during inference of $X_T$ on the task model $F$. We introduce a discriminator $D_F$ to achieve this goal. The GAN loss of feature-level alignment (FLA) is defined as:
% \begin{equation}
% \mathcal{L}_{feat}(F_f,D_{F_f}, X', X_T)=\mathbb{E}_{\mathbf{x}'\sim X'}\log D_{F_f}(F_f(\mathbf{x}'))+\mathbb{E}_{\mathbf{x}_T\sim X_T}\log [1-D_{F_f}(F_f(\mathbf{x}_T))],\\
% \label{equ:ganFeature}
% \end{equation}
\begin{equation}\small
\begin{aligned}
&\mathcal{L}_{FLA}(F_f,D_{F_f}, X', X_T)=\\
&\mathbb{E}_{\mathbf{x}'\sim X'}\log D_{F_f}(F_f(\mathbf{x}'))+\mathbb{E}_{\mathbf{x}_T\sim X_T}\log [1-D_{F_f}(F_f(\mathbf{x}_T))],\\
\end{aligned}
\label{equ:ganFeature}
\end{equation}
where $F_f(\cdot)$ is the output of the last convolution layer (\textit{i.e.} a feature map) of the encoder in $F$.

\subsection{MADAN Learning}
The proposed MADAN learning framework utilizes adaptation techniques including pixel-level alignment, cycle-consistency, dynamic semantic consistency, domain aggregation, and feature-level alignment. Combining all these components, the overall objective loss function of MADAN is:
\begin{equation}\scriptsize%\small
\begin{aligned}
&\mathcal{L}_{MADAN}(G_{S_1\rightarrow T} \cdots G_{S_M\rightarrow T}, G_{T\rightarrow S_1} \cdots G_{T\rightarrow S_M}, D_1\cdots D_M, \\
&D_T, D_A^1 \cdots D_A^M, D_{F_f}, F)=\\
&\sum_{i=1}^M\Big[\mathcal{L}_{GAN}^{S_i\rightarrow T}(G_{S_{i}\rightarrow T},D_T, X_i, X_T)+\mathcal{L}_{GAN}^{T\rightarrow S_i}(G_{T\rightarrow S_{i}},D_{i}, X_T, X_{i})\\
&+\mathcal{L}_{cyc}^{S_i\leftrightarrow T}(G_{S_i\rightarrow T},G_{T\rightarrow S_i}, X_{i}, X_T) + \mathcal{L}_{DSC}^{S_i}(G_{S_{i}\rightarrow T},X_i,F_{i}, F)\\
&+\mathcal{L}_{SAD}^{S_i}(G_{S_{1}\rightarrow T}, \dots G_{S_{i}\rightarrow T}, \dots,  G_{S_{M}\rightarrow T}, D_A^i)\\
&+\mathcal{L}_{CCD}^{S_i}(G_{T\rightarrow S_1}, \dots G_{T\rightarrow S_{i-1}}, G_{T\rightarrow S_{i+1}}, \dots, G_{T\rightarrow S_{M}}, G_{S_i\rightarrow T},D_i)\Big] \\
&+\mathcal{L}_{task}(F,X',Y) +\mathcal{L}_{FLA}(F_f,D_{F_f}, X', X_T).
\end{aligned}
\label{equ:total_loss}
\end{equation}
The training process corresponds to solving for a target model $F$ according to the optimization:
\begin{equation}\small
\begin{aligned}
F^*=\arg\min_F\min_D\max_G\mathcal{L}_{MADAN}(G, D, F),
\end{aligned}
\label{equ:optimization}
\end{equation}
where $G$ and $D$ represent all the generators and discriminators in Eq.~(\ref{equ:total_loss}), respectively.

\subsection{MADAN+ for Segmentation Adaptation}
\label{ssec:MADAN+}

There might be some problems when applying the aforementioned MADAN to pixel-wise segmentation adaptation. First, the feature-level alignment in Section~\ref{ssec:FLA} aims to align the features of the adapted images and the target images globally based on the assumption that each category's appearance frequency is identical in the adapted and target domains. This is obviously unreasonable since different categories (\textit{e.g.}, car and sky) are not uniformly distributed. Second, the image generation based on CycleGAN in Section~\ref{ssec:DAIG} only considers one crop scale. When the scale is large, local details might be missing. When it is small, the global semantics cannot be well represented. Moreover, during CycleGAN's training, a batch is composed of randomly cropped images from both the adapted and target domains at different locations. This is problematic since spatial misalignment might be caused. For example, a batch contains the upper part (\textit{e.g.} sky) in an adapted image and the lower part (\textit{e.g.} road) in a target image.

To address the above challenges, we propose (1) category-level alignment (CLA) to balance the appearance frequency of different classes, and (2) context-aware generation (CAG) using multi-scale translation and spatial alignment to generate adapted images that well preserve both global semantics and local details.

%While MADAN has achieved relatively fine results on Image Generation and feature-level Alignment (FLA), which play a big role in suppressing domain shift between source and target in global level. There're still two problems we should focus on. One is that, for different domains, each category (e.g., car, sky) is not uniformly distributed and has its unique appearance frequency. It is not plausible to assume that each category's composition is identical between source and target, thus we proposed a Category Level Alignment (CLA) module to balance the appearance frequency of different classes. On the other hand, semantic segmentation task requires a clear distinction between foreground and background. The Image Generation part of MADAN only considered only one scale size. If the value is large, it may not contain much detail information. If the value is small, the global semantic information will also be lost. Moreover, during CycleGAN's training, previous method will randomly crop image to form a batch both on source and target at different location, this will lead a problem that the batch contains the upper part sky scene in the source and the lower part road scene in the target. This misalignment of spatial information can be solved by only increasing the batch size, but on datasets like Synthia, GTAV and Cityscapes, it is hard to use a larger batch size as well as to maintain the input scale. We adopted our spatial-aligned part to solve the spatial information shift, and achieve good results under a small batch size.

\subsubsection{Category-level Alignment}
Different from the global alignment in FLA, CLA considers the alignment of local regions in different classes between the adapted and target images. Based on FLA, we can obtain the grid-wise (pseudo) labels $\aleph_{n}^{l}(\mathbf{x})$ for class $l$ of the $n$th grid in image $\mathbf{x}$. Here $l=1,\cdots,L, n=1,\cdots,N$. Following~\cite{chen2017no}, we employ one discriminator $D_C^l$ to differentiate class $l$ between the adapted and target domains.
Let $Y(\mathbf{x}_d)$ denote the labeling function for image $\mathbf{x}_d$ in domain $d$, and we have:
\begin{equation}\small
\begin{aligned}
Y(\mathbf{x}_d) =
\begin{cases}
\mathbf{y}_d,\ \ \ \ \ \ \ \ \text{if}\ \  d \in \{1,\cdots,M\},\\
F(\mathbf{x}_d),\ \ \text{if}\ \  d=T.
\end{cases}
\end{aligned}
\end{equation}
Suppose $\mathcal{R}(n)$ is the group of pixels in grid $n$, and then we can obtain the grid-wise (pseudo) labels $\aleph_{n}^{l}(\mathbf{x}_d)$ as:
\begin{equation}\small
\begin{aligned}
\aleph_{n}^{l}(\mathbf{x}_d)= \sum_{r \in \mathcal{R}(n)} \frac{|Y(\mathbf{x}_d^r) == l|}{|\mathcal{R}(n)|}.
\end{aligned}
\label{equ:Grid-wiseLabel}
\end{equation}
In order to balance the appearance frequency of the adapted and target (pseudo) labels, we normalize $\aleph_{n}^{l}(\mathbf{x}_d)$ as:
\begin{equation}\small
\begin{aligned}
\widetilde{\aleph}_{n}^{l}(\mathbf{x}_d) = \frac{\aleph_{n}^{l}(\mathbf{x}_d)}{\sum_{n=1}^N \aleph_{n}^{l}(\mathbf{x}_d)}.
\end{aligned}
\label{equ:NormalizedLabel}
\end{equation}
And then the GAN loss of CLA can be obtained as:
\begin{equation}\small
\begin{aligned}
\mathcal{L}_{CLA}(F_f,&D_C^1,\cdots,D_C^L, X', X_T)=\\
&\mathbb{E}_{\mathbf{x}'\sim X'}\sum_{l=1}^L\sum_{n=1}^N\widetilde{\aleph}_{n}^{l}(\mathbf{x}')\log D_C^l(F_f(\mathbf{x}')_n)+\\
&\mathbb{E}_{\mathbf{x}_T\sim X_T}\sum_{l=1}^L\sum_{n=1}^N\widetilde{\aleph}_{n}^{l}(\mathbf{x}_T)\log [1-D_C^l(F_f(\mathbf{x}_T)_n)].
\end{aligned}
\label{equ:CLA}
\end{equation}

\subsubsection{Context-aware Generation}
Besides global semantics, the local details of the intermediate adapted domain are more important for segmentation adaptation as compared to classification adaptation. For example, a clear boundary between the foreground and the background can contribute to the segmentation. Therefore, it is crucial to generate high-quality images during image generation process. We propose multi-scale translation and spatial alignment for the context-aware generation (CAG).

First, we resize the images from both the adapted and target domains to make the resolution aligned. Second, we randomly select a point as the center to uniformly crop both the adapted and target images into multiple sizes $\{C_1, \ldots, C_K\}$. We observe that the spatial distributions of the classes between the adapted and target domains are roughly the same (\textit{e.g.} class \textit{sky} is basically on the top of an image in both domains). Therefore, uniform cropping is crucial to ensure spatial alignment. Finally, we resize the pyramid samples into a fixed resolution. In this way, the adapted images by context-aware generation can well preserve both global semantics and local details. During reference, the full-size target image can be directly fed into the image generator to generate high-quality intermediate images.
% We observe that the classes between the adapted and target domains are spatially distributed roughly the same (\textit{e.g.} class \textit{sky} is basically on the top of an image in both domains).

Following previous steps, we can form a mini-batch $\widetilde{X}^{k}_{i}$ and $\widetilde{X}^{k}_{T}, k=1,\cdots,K$ for each scale $k$ during the training of CycleGAN. The CAG loss is defined as:
\begin{equation}\scriptsize%\small
\begin{aligned}
&\mathcal{L}_{CAG}( G_{S_1\rightarrow T} \cdots G_{S_M\rightarrow T}, G_{T\rightarrow S_1} \cdots G_{T\rightarrow S_M}, D_1\cdots D_M,D_T) = \\
& \sum_{i=1}^{M} \sum_{k=1}^{K}\Big[\mathcal{L}_{GAN}^{S_i\rightarrow T}(G_{S_{i}\rightarrow T},D_T, \widetilde{X}^{k}_i, \widetilde{X}^{k}_T)+\mathcal{L}_{GAN}^{T\rightarrow S_i}(G_{T\rightarrow S_{i}},D_{i}, \widetilde{X}^{k}_T, \widetilde{X}^{k}_i) \\
&+ \mathcal{L}_{cyc}^{S_i\leftrightarrow T}(G_{S_i\rightarrow T},G_{T\rightarrow S_i}, \widetilde{X}^{k}_i, \widetilde{X}^{k}_T) + \mathcal{L}_{DSC}^{S_i}(G_{S_{i}\rightarrow T}, \widetilde{X}^{k}_i, F_{i}, F)\Big].
\end{aligned}
\label{equ:CAG}
\end{equation}

\subsubsection{MADAN+ Learning}
Combining MADAN with CLA and CAG, we can obtain the overall objective loss function of MADAN+ as:
\begin{equation}\scriptsize%\small
\begin{aligned}
&\mathcal{L}_{MADAN+}(G_{S_1\rightarrow T} \cdots G_{S_M\rightarrow T}, G_{T\rightarrow S_1} \cdots G_{T\rightarrow S_M}, D_1\cdots D_M, \\
&D_T, D_A^1 \cdots D_A^M, D_{F_f}, F, D_C^1,\cdots,D_C^L)=\\
&\mathcal{L}_{CAG}( G_{S_1\rightarrow T} \cdots G_{S_M\rightarrow T}, G_{T\rightarrow S_1} \cdots G_{T\rightarrow S_M}, D_1\cdots D_M,D_T)\\
&+\sum_{i=1}^M\Big[\mathcal{L}_{SAD}^{S_i}(G_{S_{1}\rightarrow T}, \dots G_{S_{i}\rightarrow T}, \dots,  G_{S_{M}\rightarrow T}, D_A^i)\\
&+\mathcal{L}_{CCD}^{S_i}(G_{T\rightarrow S_1}, \dots G_{T\rightarrow S_{i-1}}, G_{T\rightarrow S_{i+1}}, \dots, G_{T\rightarrow S_{M}}, G_{S_i\rightarrow T},D_i)\Big]\\
&+\mathcal{L}_{task}(F,X',Y) +\mathcal{L}_{FLA}(F_f,D_{F_f}, X', X_T)\\
&+\mathcal{L}_{CLA}(F_f,D_C^1,\cdots,D_C^L, X', X_T).
\end{aligned}
\label{equ:MADAN+}
\end{equation}
The training process of MADAN+ is similar to MADAN.

\section{Experiments}
\label{sec:Experiments}

In this section, we first introduce the experimental settings and then compare the DA results of the proposed MADAN with several state-of-the-art approaches both quantitatively and qualitatively, followed by some empirical analysis on ablation study, feature visualization, and model interpretability. Our source code is released at: \url{https://github.com/Luodian/MADAN}.

\subsection{Experimental Settings}

In thsi section, the datasets, baselines, evaluation metrics, and implementation details are described.

\subsubsection{Datasets}.

\textbf{Digit Recognition.} Digits-five includes 5 digit image datasets sampled from different domains, including \emph{handwritten} \textbf{mt} (MNIST) \cite{lecun1998gradient}, \emph{combined} \textbf{mm} (MNIST-M) \cite{ganin2015unsupervised}, \emph{street image} \textbf{sv} (SVHN) \cite{netzer2011reading}, \emph{synthetic} \textbf{sy} (Synthetic Digits) \cite{ganin2015unsupervised}, and \emph{handwritten} \textbf{up} (USPS) \cite{hull1994database}. Following~\cite{xu2018deep,peng2019moment}, we sample 25,000 images for training and 9,000 for testing in \textbf{mt}, \textbf{mm}, \textbf{sv}, \textbf{sy}, and select the entire 9,298 images in \textbf{up} as a domain.

\textbf{Object Classification.} Office-31~\cite{saenko2010adapting} contains 4,110 images within 31 categories, which are collected from office environment in three image domains: \textbf{A} (Amazon) downloaded from amazon.com, \textbf{W} (Webcam) and \textbf{D} (DSLR) taken by web camera and digital SLR camera, respectively.

Office+Caltech-10~\cite{gong2013connecting} consists of the 10 overlapping categories shared by Office-31~\cite{saenko2010adapting} and \textbf{C} (Caltech-256)~\cite{griffin2007caltech}. Totally there are 2,533 images.

Office-Home~\cite{venkateswara2017deep} is a larger object dataset with 30,475 images within 65 categories. There are 4 different domains: Artistic images (\textbf{Ar}), Clip-Art images (\textbf{Cl}), Product images (\textbf{Pr}) and Real-World images (\textbf{Rw}).

\begin{table}[!t]
\centering\small%\scriptsize
%\caption{Classification accuracy (\%) on Digits-five dataset for multi-source unsupervised domain adaptation. The best method trained on the source domains is emphasized in bold and red. Our method achieves accuracy, significantly outperforming the state-of-the-arts.}
\caption{Comparison with the state-of-the-art DA methods for digit recognition on Digits-five dataset measured by classification accuracy (\%). The best method is emphasized in bold.}
\resizebox{\linewidth}{!}{%
\begin{tabular}
{c | c | c  c  c  c  c | c}
\toprule
Standard & Method & mm & mt & up & sv & sy & Avg\\
\hline
\multirow{2}{1.7cm}{\centering Source-only}   &  Combined & 63.7  &  92.3 & 87.2  &  66.3  &  84.8  &  78.9\\
 & Single-best  & 59.2  & 97.2  & 84.7  &  77.7 & 85.2 & 80.8  \\
\hline
\multirow{5}{1.7cm}{\centering Single-best DA}   & DAN~\cite{long2015learning}  & 63.8  &  96.3 & 94.2  & 62.5 & 85.4  &  80.4\\
   & CORAL~\cite{sun2016return}  & 62.5  & 97.2  & 93.5  & 64.4 & 82.8  & 80.1 \\
   & DANN~\cite{ganin2016domain}  & 71.3  & 97.6  & 92.3  & 63.5  &  85.3 & 82.0 \\
   & ADDA~\cite{tzeng2017adversarial}  & 71.6  & 97.9  & 92.8  & 75.5 & 86.5  & 84.9 \\
   & CyCADA~\cite{hoffman2018cycada} & 72.4	& 98.0 & 92.4 & 76.7 & 87.4 & 85.4 \\
%   & ADDA WGAN~\cite{tzeng2017adversarial}  & 88.9  & 98.8  & 91.6  & 79.5 & 89.6  & 89.7 \\
\hline
\multirow{4}{1.7cm}{\centering Source-combined DA}   & DAN~\cite{long2015learning}  &  67.9 & 97.5  & 93.5  &  67.8 & 86.9  & 82.7 \\
   & DANN~\cite{ganin2016domain}  & 70.8  & 97.9  &  93.5 &  68.5 & 87.4  & 83.6 \\
   & ADDA~\cite{tzeng2017adversarial}  & 72.3  & 97.9  & 93.1  &  75.0 & 86.7  &  85.0 \\
    & CyCADA~\cite{hoffman2018cycada} & 72.4 & 98.1 & 93.1 & 75.2 & 	86.9 & 85.1 \\
%   & ADDA (ours) & 58.5  & 91.1  & 89.5  &  70.8 & 77.8  &  77.5 \\
\hline
\multirow{5}{1.7cm}{\centering Multi-source DA}   & DCTN~\cite{xu2018deep}  & 70.5  & 96.2  & 92.8  & 77.6  & 86.8  &  84.8 \\
   & MDAN~\cite{zhao2018adversarial}  & 69.5  & 98.0  &  92.5  & 69.2  & 87.4  & 83.3 \\
   & M3SDA~\cite{peng2019moment} & 72.8 & 98.6 & 96.1 & \textbf{81.3} & 89.6 & 87.7\\
   & MDDA~\cite{zhao2020multi}  & 78.6  & 98.8  &  93.9  &  79.3 &  89.7 &  88.1\\
   & \textbf{MADAN (ours)} & \textbf{82.9} & \textbf{99.7} & \textbf{96.7} & 80.2 & \textbf{95.2} & \textbf{90.9} \\
\bottomrule
%Oracle   & AlexNet  &   &   &   &  &  & \\
%\hline
\end{tabular}
}
\label{tab:DigitsFive}
\end{table}

\begin{table}[!t]
\centering\footnotesize%\scriptsize%\small
\caption{Comparison with the state-of-the-art DA methods for object classification on Office31 dataset measured by classification accuracy (\%). The best method is emphasized in bold.}%significantly outperforming the state-of-the-art approaches.
\resizebox{0.85\linewidth}{!}{%
\begin{tabular}
{c | c | c  c  c | c }
\toprule
%\multirow{2}{*}{Standards} & \multirow{2}{*}{Models} & A,W & A,D & D,W & \multirow{2}{*}{\ Avg\ }\\
%   &   &  $\rightarrow$D & $\rightarrow$W  & $\rightarrow$A  & \\
Standard & Method & D & W & A & Avg\\
\hline
\multirow{2}{1.7cm}{\centering Source-only}   &  Combined & 97.1  & 92.0  & 51.6  & 80.2 \\
 & Single-best  & 99.0  & 95.3  & 50.2  &  81.5 \\
\hline
\multirow{9}{1.7cm}{\centering Single-best DA}   & TCA~\cite{pan2011domain}  &  95.2 &  93.2 &  51.6 & 80.0 \\
& GFK~\cite{gong2012geodesic}  & 95.0  &  95.6 &  52.4 &  81.0 \\
& DDC~\cite{tzeng2015simultaneous}  & 98.5  & 95.0  &  52.2 & 81.9 \\
& DRCN~\cite{ghifary2016deep}  & 99.0  & 96.4  &  56.0 & 83.8 \\
& RevGrad~\cite{ganin2015unsupervised}  & 99.2  & 96.4  &  53.4 & 83.0 \\
& DAN~\cite{long2015learning}  & 99.0  & 96.0  & 54.0  & 83.0 \\
& RTN~\cite{long2016unsupervised}  & \textbf{99.6}  & 96.8  & 51.0  & 82.5 \\
& ADDA~\cite{tzeng2017adversarial}  & 99.4  & 95.3  & 54.6  & 83.1 \\
& CyCADA~\cite{hoffman2018cycada} &  98.9 & 94.8 & 53.2 & 82.3 \\
%& ADDA(ours) & 99.2  & 96.0  & 54.6  & 83.2 \\
\hline
\multirow{4}{1.7cm}{\centering Source-combined DA}   & RevGrad~\cite{ganin2015unsupervised}  & 98.8  & 96.2  &  54.6 & 83.2 \\
   & DAN~\cite{long2015learning}  & 98.8  & 96.2  & 54.9  &  83.3 \\
   & ADDA~\cite{tzeng2017adversarial}  & 99.2  &  96.0 & 55.9  & 83.7  \\
   & CyCADA~\cite{hoffman2018cycada} & 99.0 & 96.2 & 54.2 & 83.1 \\
%   & ADDA(ours) & 98.2  & 96.9  & 52.6  & 82.6 \\
\hline
\multirow{4}{1.7cm}{\centering Multi-source DA}   & DCTN~\cite{xu2018deep}  & \textbf{99.6}  &  96.9 & 54.9  & 83.8  \\
   & MDAN~\cite{zhao2018adversarial}  & 99.2  & 95.4  &  55.2 &  83.3  \\
   & MDDA~\cite{zhao2020multi}  & 99.2  & 97.1  & 56.2  &  84.2  \\
   & \textbf{MADAN (ours)} & 99.4 & \textbf{98.4} & \textbf{63.9} & \textbf{87.2} \\
\bottomrule
%Oracle   & AlexNet  &   &   &   &   \\
%\hline
\end{tabular}
}
\label{tab:Office31}
\end{table}

\textbf{Semantic Segmentation.} Cityscapes~\cite{cordts2016cityscapes} contains vehicle-centric urban street images collected from a moving vehicle in 50 cities from Germany and neighboring countries. There are 5,000 images with pixel-wise annotations. The images have resolution of $2048 \times 1024$ and are labeled into 19 classes.

BDDS~\cite{yu2018bdd100k} contains 10,000 real-world dash cam video frames with accurate pixel-wise annotations.  It has a compatible label space with Cityscapes and the image resolution is $1280 \times 720$..

GTA~\cite{richter2016playing} is a vehicle-egocentric image dataset collected in the high-fidelity rendered computer game GTA-V. It contains 24,966 images (video frames)  with the resolution $1914\times1052$. There are 19 classes compatible with Cityscapes.

SYNTHIA~\cite{ros2016synthia} is a large synthetic dataset. To pair with Cityscapes, a subset, named SYNTHIA-RANDCITYSCAPES, is designed with 9,400 images with resolution $960\times720$ which are automatically annotated with 16 object classes, one void class, and some unnamed classes.

\subsubsection{Baselines}

We compare  MADAN with the following methods. \textbf{(1) Source-only}, \textit{i.e.} train on the source domains and directly test on the target domain. We can view this as a lower bound of DA. \textbf{(2) Single-source DA}, perform multi-source DA via single-source DA. \textbf{(3) Multi-source DA}, extend some single-source DA method to multi-source settings.

\begin{table}[!t]
\centering\footnotesize%\scriptsize%\small
\caption{Comparison with the state-of-the-art DA methods for object classification on Office+Caltech-10 dataset measured by classification accuracy (\%). The best method is emphasized in bold.}%significantly outperforming the state-of-the-art approaches.
\resizebox{\linewidth}{!}{%
\begin{tabular}
{c | c | c  c  c  c | c }
\toprule
%\multirow{2}{*}{Standards} & \multirow{2}{*}{Models} & A,W & A,D & D,W & \multirow{2}{*}{\ Avg\ }\\
%   &   &  $\rightarrow$D & $\rightarrow$W  & $\rightarrow$A  & \\
Standard & Method & W & D & C & A & Avg\\
\hline
\multirow{2}{1.7cm}{\centering Source-only}   &  Combined & 93.1 & 98.4 & 81.9	 & 93.1 & 91.6 \\
 & Single-best  & 98.9 & 99.2 & 82.5 & 91.2 & 93.0 \\
\hline
\multirow{2}{1.7cm}{\centering Single-best DA} & ADDA~\cite{tzeng2017adversarial}  & 99.1 & 98.0 & 88.8 & 94.5 & 95.1\\
& CyCADA~\cite{hoffman2018cycada} &  98.9 & 97.3 & 89.7 & 96.2 & 95.5\\
%& ADDA(ours) & 99.2  & 96.0  & 54.6  & 83.2 \\
\hline
\multirow{3}{1.7cm}{\centering Source-combined DA}    & DAN~\cite{long2015learning}  & 99.3  & 98.2  & 89.7  & 94.8  & 95.5\\
   & ADDA~\cite{tzeng2017adversarial}  & 99.4  & 98.2  & 90.2  & 95.0 & 95.7\\
   & CyCADA~\cite{hoffman2018cycada} & 99.0  & 97.8  & 91.0  & 95.9  & 95.9 \\
\hline
\multirow{4}{1.7cm}{\centering Multi-source DA}   & DCTN~\cite{xu2018deep}  & 99.4  & 99.0  & 90.2  & 92.7  & 95.3 \\
   & MDAN~\cite{zhao2018adversarial}  & 98.1  & 98.2  & 89.5  & 92.2  & 94.5 \\
   & M3SDA~\cite{peng2019moment}  & \textbf{99.5}  & 99.2  & 92.2  & 94.5  & 96.4 \\
   & \textbf{MADAN (ours)} & 99.2  & \textbf{100.0}  & \textbf{97.2}  & \textbf{97.9}  & \textbf{98.6} \\
\bottomrule
%Oracle   & AlexNet  &   &   &   &   \\
%\hline
\end{tabular}
}
\label{tab:Office-Caltech}
\end{table}

\begin{table}[!t]
\centering\footnotesize%\scriptsize%\small
\caption{Comparison with the state-of-the-art DA methods for object classification on Office-Home dataset measured by classification accuracy (\%). The best method is emphasized in bold.}%significantly outperforming the state-of-the-art approaches.
\resizebox{\linewidth}{!}{%
\begin{tabular}
{c | c | c  c  c  c | c }
\toprule
%\multirow{2}{*}{Standards} & \multirow{2}{*}{Models} & A,W & A,D & D,W & \multirow{2}{*}{\ Avg\ }\\
%   &   &  $\rightarrow$D & $\rightarrow$W  & $\rightarrow$A  & \\
Standard & Method & Rw & Pr & Cl & Ar & Avg\\
\hline
\multirow{2}{2.2cm}{\centering Source-only}   &  Combined & 68.1 & 76.9 & 48.9	 & 65.4 & 64.8 \\
 & Single-best  & 60.4 & 59.9 & 41.2 & 53.9 & 53.9\\
\hline
\multirow{4}{2.2cm}{\centering Single-best \\DA} & DAN~\cite{long2015learning} & 67.9 & 74.3 & 51.5 & 63.1 & 64.2\\
& DANN~\cite{ganin2016domain}  & 70.1 & 76.8 & 51.8 & 63.2 & 65.5\\
& JAN~\cite{long2017deep} & 68.9 & 76.8 & 52.4 & 63.9 & 65.5\\
& CyCADA~\cite{hoffman2018cycada} & 77.4  & 75.3  & 51.9  & 68.7  & 68.3 \\
\hline
\scriptsize{\multirow{1}{2.2cm}{\centering Source-combined DA}}    & CyCADA~\cite{hoffman2018cycada} & 79.4  &  72.9  & 50.4  & 62.6  &  66.3 \\
\hline
\multirow{2}{2.2cm}{\centering Multi-source \\DA}    & MDAN~\cite{zhao2018adversarial}  & 76.3  & 69.2  & 49.7  & 64.9  & 65.0 \\
   & \textbf{MADAN (ours)} & \textbf{81.5}  & \textbf{78.2}  & \textbf{54.9}  & \textbf{66.8}  & \textbf{70.4} \\
\bottomrule
%Oracle   & AlexNet  &   &   &   &   \\
%\hline
\end{tabular}
}
\label{tab:Office-Home}
\end{table}

\begin{table*}[]
\caption{Comparison with the state-of-the-art DA methods for semantic segmentation from GTA and SYNTHIA to Cityscapes using FCN-VGG16 backbone. The best class-wise IoU and mIoU trained on the source domains are emphasized in bold (similar below).}
\resizebox{\textwidth}{!}{%
\begin{tabular}{c|c|cccccccccccccccc|c}
\toprule
\multirow{1}{*}[1.8em]{Standard}&\multirow{1}{*}[1.8em]{Method}  & \rot{road} & \rot{sidewalk} & \rot{building} & \rot{wall} & \rot{fence} & \rot{pole} & \rot{t-light} & \rot{t-sign} & \rot{vegettion} & \rot{sky}   & \rot{person} & \rot{rider} & \rot{car} & \rot{bus} & \rot{m-bike} & \rot{bicycle} & \rot{mIoU} \\ \hline
\multirow{3}{*}{Source-only}&GTA & 54.1 & 19.6 & 47.4 & 3.3 & 5.2 & 3.3 & 0.5 & 3.0 & 69.2 & 43.0 & 31.3 & 0.1 & 59.3 & 8.3 & 0.2 & 0.0 & 21.7  \\
&SYNTHIA & 3.9 & 14.5 & 45.0 & 0.7 & 0.0 & 14.6 & 0.7 & 2.6 & 68.2 & 68.4 & 31.5 & 4.6 & 31.5 & 7.4 & 0.3 & 1.4 & 18.5 \\
&GTA+SYNTHIA & 44.0 & 19.0 & 60.1 & 11.1 & 13.7 & 10.1 & 5.0 & 4.7 & 74.7 & 65.3 & 40.8 & 2.3 & 43.0 & 15.9 & 1.3 & 1.4 & 25.8 \\ \hline
\multirow{6}{*}{GTA-only DA}&FCN Wld~\cite{hoffman2016fcns} & 70.4 & 32.4 & 62.1 & 14.9 & 5.4 & 10.9 & 14.2 & 2.7 & 79.2 & 64.6 & 44.1 & 4.2 & 70.4 & 7.3 & 3.5 & 0.0 & 27.1 \\
&CDA~\cite{zhang2017curriculum} & 74.8 & 22.0 & 71.7 & 6.0 & 11.9 & 8.4 & 16.3 & 11.1 & 75.7 & 66.5 & 38.0 & 9.3 & 55.2 & 18.9 & 16.8 & 14.6 & 28.9 \\
&ROAD~\cite{chen2018road} & 85.4 & 31.2 & 78.6 & \textbf{27.9} & \textbf{22.2} & 21.9 & 23.7 & 11.4 & 80.7 & 68.9 & 48.5 & 14.1 & 78.0 & 23.8 & 8.3 & 0.0 & 39.0 \\
&AdaptSeg~\cite{tsai2018learning} & 87.3 &29.8 &78.6 &21.1 &18.2 &22.5 &21.5 &11.0 &79.7  &71.3 &46.8 &6.5 &\textbf{80.1} &26.9 &10.6 &0.3 &38.3 \\
&CyCADA~\cite{hoffman2018cycada} & 85.2 & 37.2 & 76.5 & 21.8 & 15.0 & 23.8 & 22.9 & 21.5 & 80.5 & 60.7 & 50.5 & 9.0 & 76.9 & 28.2 & 4.5 & 0.0 & 38.7 \\
&DCAN~\cite{wu2018dcan} &82.3 &26.7 &77.4 &23.7 &20.5 &20.4 &\textbf{30.3} &15.9 &\textbf{80.9} &69.5 & 52.6 &11.1 &79.6 &21.2 &17.0 &6.7 &39.8 \\
\hline
\multirow{5}{*}{SYNTHIA-only DA}&FCN Wld~\cite{hoffman2016fcns} & 11.5 & 19.6 & 30.8 & 4.4 & 0.0 & 20.3 & 0.1 & 11.7 & 42.3 & 68.7 & 51.2 & 3.8 & 54.0 & 3.2 & 0.2 & 0.6 & 20.2 \\
&CDA~\cite{zhang2017curriculum} & 65.2 & 26.1 & 74.9 & 0.1 & 0.5 & 10.7 & 3.7 & 3.0 & 76.1 & 70.6 & 47.1 & 8.2 & 43.2 & 20.7 & 0.7 & 13.1 & 29.0 \\
&ROAD~\cite{chen2018road} & 77.7 & 30.0 & 77.5 & 9.6 & 0.3 & 25.8 & 10.3 & 15.6 & 77.6 & 79.8 & 44.5 & 16.6 & 67.8 & 14.5 & 7.0 & 23.8 & 36.2 \\
&CyCADA~\cite{hoffman2018cycada} & 66.2 & 29.6 & 65.3 & 0.5 & 0.2 & 15.1 & 4.5 & 6.9 & 67.1 & 68.2 & 42.8 & 14.1 & 51.2 & 12.6 & 2.4 & 20.7 & 29.2 \\
&DCAN~\cite{wu2018dcan} &79.9 &30.4 &70.8 &1.6 &0.6 &22.3 &6.7 &\textbf{23.0} &76.9 &73.9 &41.9 &16.7 &61.7 &11.5 &10.3 &\textbf{38.6 }&35.4 \\ \hline
\multirow{1}{*}{Source-combined DA}&CyCADA~\cite{hoffman2018cycada} & 82.8 & 35.8 & 78.2 & 17.5 & 15.1 & 10.8 & 6.1 & 19.4 & 78.6 & 77.2 & 44.5 & 15.3 & 74.9 & 17.0 & 10.3 & 12.9 & 37.3 \\ \hline
&MDAN~\cite{zhao2018adversarial} & 64.2 & 19.7 & 63.8 & 13.1 & 19.4 & 5.5 & 5.2 & 6.8 & 71.6 & 61.1 & 42.0 & 12.0 & 62.7 & 2.9 & 12.3 & 8.1 & 29.4 \\
\multirow{-1}{*}{Multi-source DA}&\textbf{MADAN~(Ours)} & 86.2 & 37.7 & \textbf{79.1} & 20.1 & 17.8 & 15.5 & 14.5 & 21.4 & 78.5 & 73.4 & 49.7 & 16.8 & 77.8 & 28.3 & \textbf{17.7} & 27.5 & 41.4  \\
& \textbf{MADAN+~(Ours)} & \textbf{87.9} & \textbf{41.0} & 76.4 & 21.4 & 1.3 & \textbf{28.4} & 20.3 & 22.3 & 77.3 & \textbf{80.0} & \textbf{54.9} & \textbf{21.5} & \textbf{80.1} & \textbf{29.7} & 15.1 & 26.5 & \textbf{42.8} \\
\hline
Oracle-Train on Target & FCN~\cite{long2015fully} & 96.4 & 74.5 & 87.1 & 35.3 & 37.8 & 36.4 & 46.9 & 60.1 & 89.0 & 89.8 & 65.6 & 35.9 & 76.9 & 64.1 & 40.5 & 65.1 & 62.6  \\
\bottomrule
\end{tabular}
}
\label{tab:comparison_with_SOTA_CS_FCN}
\end{table*}

\begin{table*}[]
\caption{Comparison with the state-of-the-art DA methods for semantic segmentation from GTA and SYNTHIA to BDDS using FCN-VGG16 backbone.}
\resizebox{\textwidth}{!}{%
\begin{tabular}{c|c|cccccccccccccccc|c}
\toprule
\multirow{1}{*}[1.8em]{Standard}&\multirow{1}{*}[1.8em]{Method}  & \rot{road} & \rot{sidewalk} & \rot{building} & \rot{wall} & \rot{fence} & \rot{pole} & \rot{t-light} & \rot{t-sign} & \rot{vegettion} & \rot{sky}   & \rot{person} & \rot{rider} & \rot{car} & \rot{bus} & \rot{m-bike} & \rot{bicycle} & \rot{mIoU} \\ \hline
\multirow{3}{*}{Source-only}&GTA & 50.2 & 18.0 & 55.1 & 3.1 & 7.8 & 7.0 & 0.0 & 3.5 & 61.0 & 50.4 & 19.2 & 0.0 & 58.1 & 3.2 & \textbf{19.8} & 0.0 & 22.3  \\
&SYNTHIA & 7.0 & 6.0 & 50.5 & 0.0 & 0.0 & 15.1 & 0.2 & 2.4 & 60.3 & \textbf{85.6} & 16.5 & 0.5 & 36.7 & 3.3 & 0.0 & 3.5 & 17.1 \\
&GTA+SYNTHIA & 54.5 & 19.6 & 64.0 & 3.2 & 3.6 & 5.2 & 0.0 & 0.0 & 61.3 & 82.2 & 13.9 & 0.0 & 55.5 & 16.7 & 13.4 & 0.0 & 24.6 \\ \hline
GTA-only DA & CyCADA~\cite{hoffman2018cycada} &\textbf{77.9} & 26.8 & 68.8 & 13.0 & 19.7 & 13.5 & 18.2 & \textbf{22.3} & 64.2 & 84.2 & 39.0 & \textbf{22.6} & 72.0 & 11.5 & 15.9 & 2.0 & 35.7 \\
\hline
SYNTHIA-only DA & CyCADA~\cite{hoffman2018cycada} &55.0 & 13.8 & 45.2 & 0.1 & 0.0 & 13.2 & 0.5 & 10.6 & 63.3 & 67.4 & 22.0 & 6.9 & 52.5 & 10.5 & 10.4 & 13.3  & 24.0 \\
\hline
Source-combined DA & CyCADA~\cite{hoffman2018cycada} &61.5 & 27.6 & 72.1 & 6.5 & 2.8 & 15.7 & 10.8 & 18.1 & 78.3 & 73.8 & 44.9 & 16.3 & 41.5 & 21.1 & 21.8 & \textbf{25.9} & 33.7 \\
\hline
&MDAN~\cite{zhao2018adversarial} & 35.9 & 15.8 & 56.9 & 5.8 & 16.3 & 9.5 & 8.6 & 6.2 & 59.1 & 80.1 & 24.5 & 9.9 & 53.8 & 11.8 & 2.9 & 1.6 & 25.0 \\
\multirow{-1}{*}{Multi-source DA}&\textbf{MADAN~(Ours)} & 60.2 & 29.5 & 66.6 & 16.9 & 10.0 & 16.6 & 10.9 & 16.4 & 78.8 & 75.1 & 47.5 & 17.3 & 48.0 & \textbf{24.0} & 13.2 & 17.3 & 36.3  \\
&\textbf{MADAN+~(Ours)} &  75.2 & \textbf{29.8} & \textbf{83.3} & \textbf{27.2} & \textbf{20.7} & \textbf{37.8} & \textbf{23.2} & 20.6 & \textbf{81.1} & 83.5 & \textbf{50.1} & 9.8 & \textbf{80.2} & 13.2 & 11.6 & 18.1 & \textbf{41.6} \\
\hline
Oracle-Train on Target & FCN~\cite{long2015fully} & 91.7 & 54.7 & 79.5 & 25.9 & 42.0 & 23.6 & 30.9 & 34.6 & 81.2 & 91.6 & 49.6 & 23.5 & 85.4 & 64.2 & 28.4 & 41.1 & 53.0  \\
\bottomrule
\end{tabular}
}
\label{tab:comparison_with_SOTA_BDD_FCN}
\end{table*}

For digit recognition and object classification, we employ two strategies to implement the source-only and single-source DA standards: (1) single-best, \textit{i.e.} performing adaptation on each single source and selecting the best adaptation result in the target test set; (2) source-combined, \textit{i.e.} all source domains are combined into a traditional single source. The compared single-source DA includes TCA~\cite{pan2011domain}, GFK~\cite{gong2012geodesic}, DDC~\cite{tzeng2015simultaneous}, DRCN~\cite{ghifary2016deep}, RevGrad~\cite{ganin2015unsupervised}, DAN~\cite{long2015learning}, RTN~\cite{long2016unsupervised}, CORAL~\cite{sun2016return}, DANN~\cite{ganin2016domain},  ADDA~\cite{tzeng2017adversarial}, JAN~\cite{long2017deep}, and CyCADA~\cite{hoffman2018cycada}. The compared multi-source DA includes DCTN~\cite{xu2018deep}, MDAN~\cite{zhao2018adversarial}, M3SDA~\cite{peng2019moment}, and MDDA~\cite{zhao2020multi}. Please note that we only compare the methods that report the results on corresponding tasks.

For semantic segmentation, besides source combined, we also implement the source-only and single-source DA standards on each source, \textit{i.e.} performing adaptation on each single source. The compared single-source DA includes FCNs Wld~\cite{hoffman2016fcns}, CDA~\cite{zhang2017curriculum}, ROAD~\cite{chen2018road}, AdaptSeg~\cite{tsai2018learning}, CyCADA~\cite{hoffman2018cycada}, and DCAN~\cite{wu2018dcan}. Since MADAN is the first work on MDA for segmentation, we extend the original classification network in MDAN to our segmentation task for comparison.
We also report the results of an oracle setting, where the segmentation model is both trained and tested on the target domain.

%We compare  MADAN with the following methods. \textbf{(1) Source-only}, \textit{i.e.} train on the source domains and test on the target domain directly. We can view this as a lower bound of DA. \textbf{(2) Single-source DA}, perform multi-source DA via single-source DA, including FCNs Wld~\cite{hoffman2016fcns}, CDA~\cite{zhang2017curriculum}, ROAD~\cite{chen2018road}, AdaptSeg~\cite{tsai2018learning}, CyCADA~\cite{hoffman2018cycada}, and DCAN~\cite{wu2018dcan}. \textbf{(3) Multi-source DA}, extend some single-source DA method to multi-source settings, including MDAN~\cite{zhao2018adversarial}. For comparison, we also report the results of an oracle setting, where the segmentation model is both trained and tested on the target domain.
%For the source-only and single-source DA standards, we employ two strategies: (1) single-source, \textit{i.e.} performing adaptation on each single source; (2) source-combined, \textit{i.e.} all source domains are combined into a traditional single source. For MDAN, we extend the original classification network for our segmentation task.

\subsubsection{Evaluation Metric}

For classification (digit recognition and object classification) adaptation, we employ the average classification accuracy of all categories to evaluate the results following~\cite{ganin2016domain,tzeng2017adversarial,hoffman2018cycada}. The larger the classification accuracy is, the better the result is.

For pixel-wise segmentation adaptation, we employ class-wise intersection-over-union (cwIoU) and mean IoU (mIoU) to evaluate the results of each class and all classes as in~\cite{hoffman2016fcns,zhang2017curriculum,hoffman2018cycada}. Let $\mathcal{P}_l$ and $\mathcal{G}_l$ respectively denote the predicted and ground-truth pixels that belong to class $l$, and then $\displaystyle cwIoU_l = \frac{|\mathcal{P}_l \cap \mathcal{G}_l|}{|\mathcal{P}_l \cup \mathcal{G}_l|}$, $\displaystyle mIoU=\frac{1}{L}\sum\nolimits_{l=1}^{L}cwIoU_l$, where $|\cdot|$ denotes the cardinality of a set. Larger cwIoU and mIoU values represent better performances.

\subsubsection{Implementation Details}

Although MADAN can be trained in an end-to-end manner, due to constrained  hardware resources, we train it in three stages. First, we train several CycleGANs (9 residual blocks for generator and 4 convolution layers for discriminator)~\cite{zhu2017unpaired} without semantic consistency loss for each source and target pair, and then train a task model $F$ on the adapted images with corresponding labels from the source domains. Second, after updating $F_A$ with $F$ trained above, we generate adapted images using CycleGAN with the proposed DSC loss in Eq.~(\ref{equ:semConsisSi}) and aggregate different adapted domains using SAD and CCD. Finally, we train the task model $F$ on the newly adapted images in the aggregated domain with feature-level alignment. The above stages are trained iteratively.
We leave the end-to-end training as future work by deploying model parallelism or experimenting with larger GPU memory.

% In Digits-five experiments, we use three convlutional layers and two fully connected layers as encoder and one fully connected layer as classifier. We use a stochastic gradient descent (SGD) optimizer with a learning rate of 0.001, batch size of 128 in source classifier pretraining; while we use an adaptive moment estimation (ADAM) optimizer with a learning rate of 0.0001, batch size of 64 for all the components in adversarial discriminative adaptation.

\begin{table*}[]
\caption{Comparison with the state-of-the-art DA methods for semantic segmentation from GTA and SYNTHIA to Cityscapes using DeepLabV2-ResNet101 backbone. The best class-wise IoU and mIoU trained on the source domains are emphasized in bold (similar below).}
\resizebox{\textwidth}{!}{%
\begin{tabular}{c|c|cccccccccccccccc|c}
\toprule
\multirow{1}{*}[1.8em]{Standard}&\multirow{1}{*}[1.8em]{Method}  & \rot{road} & \rot{sidewalk} & \rot{building} & \rot{wall} & \rot{fence} & \rot{pole} & \rot{t-light} & \rot{t-sign} & \rot{vegettion} & \rot{sky}   & \rot{person} & \rot{rider} & \rot{car} & \rot{bus} & \rot{m-bike} & \rot{bicycle} & \rot{mIoU} \\ \hline
\multirow{3}{*}{Source-only}&GTA & 74.2 & 27.5 & 69.9 & 10.5 & 8.7 & 23.0 & 0.2 & 0.2 & 77.9 & 78.6 & 45.3 & 12.3 & 74.6 & 26.1 & 16.2 & 28.5 & 35.9  \\
&SYNTHIA & 40.3 & 19.5 & 57.6 & 6.6 & 0.1 & 30.1 & 3.4 & 15.1 & 76.8 & 76.9 & 50.9 & 8.4 & 72.9 & 30.0 & 9.7 & 16.2 & 32.2 \\
&GTA+SYNTHIA & 77.1 & 32.4 & 75.3 & 13.8 & 11.5 & 29.0 & 13.7 & 10.3 & 81.5 & 79.1 & 53.1 & 10.2 & 80.2 & 39.0 & 21.9 & 11.5 & 40.0 \\ \hline
\multirow{4}{*}{GTA-only DA} & AdaptSeg~\cite{tsai2018learning} & 86.5 & 25.9 & 79.8 & 22.1 & 20.0 & 23.6 & 33.1 & 21.8 & 81.8 & 75.9 & 57.3 & 26.2 & 76.3 & 32.1 & 29.5 & 32.5 & 41.4 \\
& DCAN~\cite{wu2018dcan} & 85.0 & 30.8 & \textbf{81.3} & 25.8 & 21.2 & 22.2 & 25.4 & 26.6 & 83.4 & 76.2 & 58.9 & 24.9 & 80.7 & 42.9 & 26.9 & 11.6	& 41.7 \\
& CyCADA~\cite{hoffman2018cycada} & 86.7 & 35.6 & 80.1 & 19.8 & 17.5 & 38.0 & 39.9 & \textbf{41.5} & 82.7 & 73.6 & \textbf{64.9} & 19.0 & 65.0 & 28.6 & 31.1 & \textbf{42.0} & 47.9 \\
& CLAN~\cite{luo2019taking} & 87.0 & 27.1 & 79.6 & \textbf{27.3} & \textbf{23.3} & 28.3 & 35.5 & 24.2 & \textbf{83.6} & 74.2 & 58.6 & \textbf{28.0} & 76.2 & 36.7 & \textbf{31.9} & 31.4 & 47.1 \\
\hline
\multirow{1}{*}{SYNTHIA-only DA} & CyCADA~\cite{hoffman2018cycada} & 82.9 & 39.0 & 79.5 & 21.2 & 4.7 & 29.5 & 13.2 & 11.7 & 78.3 & 75.8 & 53.3 & 13.7 & 83.8 & 40.0 & 20.6 & 24.4 & 42.0 \\ \hline
\multirow{1}{*}{Source-combined DA} & CyCADA~\cite{hoffman2018cycada} &   86.8 & 41.4 & 74.7 & 15.5 & 3.4 & 27.3 & 3.8 & 0.2 & 73.2 & 72.4 & 51.9 & 12.7 & 82.7 & 41.8 & 18.5 & 23.3 & 39.3 \\ \hline
\multirow{3}{*}{Multi-source DA}&MDAN~\cite{zhao2018adversarial} & 80.6 & 34.4 & 73.9 & 15.9 & 1.9 & 22.9 & 0.1 & 0.0 & 73.6 & 58.9 & 48.4 & 12.2 & 78.8 & 36.8 & 14.2 & 23.7 & 36.0 \\
&\textbf{MADAN~(Ours)} & 88.1 & 46.1 & 79.9 & 26.4 & 7.4 & 30.6 & 19.0 & 19.9 & 80.4 & 75.9 & 55.6 & 15.6 & 84.1 & \textbf{47.0} & 23.3 & 26.3 & 45.4  \\
& \textbf{MADAN+~(Ours)} & \textbf{90.9} & \textbf{49.7} & 64.9 & 24.6 & 13.0 & \textbf{39.2} & \textbf{40.0} & 21.4 & 80.2 & \textbf{86.1} & 57.3 & 25.0 & \textbf{84.7} & 35.7 & 25.2 & 38.2 & \textbf{48.5} \\
\hline
Oracle-Train on Target & DeepLabV2~\cite{chen2017deeplab} & 97.1 & 78.7 & 89.4 & 52.0 & 49.7 & 39.9 & 26.9 & 47.1 & 89.1 & 89.8 & 64.6 & 29.2 & 90.4 & 78.0 & 41.4 & 65.3 &	64.2  \\
\bottomrule
\end{tabular}
}
\label{tab:comparison_with_SOTA_CS_Deeplab}
\end{table*}

\begin{table*}[]
\caption{Comparison with the state-of-the-art DA methods for semantic segmentation from GTA and SYNTHIA to BDDS using DeepLabV2-ResNet101 backbone.}
\resizebox{\textwidth}{!}{%
\begin{tabular}{c|c|cccccccccccccccc|c}
\toprule
\multirow{1}{*}[1.8em]{Standard}&\multirow{1}{*}[1.8em]{Method}  & \rot{road} & \rot{sidewalk} & \rot{building} & \rot{wall} & \rot{fence} & \rot{pole} & \rot{t-light} & \rot{t-sign} & \rot{vegettion} & \rot{sky}   & \rot{person} & \rot{rider} & \rot{car} & \rot{bus} & \rot{m-bike} & \rot{bicycle} & \rot{mIoU} \\ \hline
\multirow{3}{*}{Source-only}&GTA & 57.4 & 17.3 & 61.8 & 5.6 & 15.1 & 27.4 & \textbf{28.6} & 15.8 & 61.2 & 82.3 & 47.7 & 5.4 & 72.2 & 28.9 & \textbf{29.7} & 1.2 &	34.9  \\
&SYNTHIA & 14.9 & 10.8 & 47.2 & 0.5 & 0.0 & 23.8 & 0.4 & 3.5 & 67.8 & \textbf{85.6} & 32.4 & 14.4 & 69.5 & 28.2 & 12.7 & 8.1 &	26.2 \\
&GTA+SYNTHIA & 55.3 & 20.9 & 73.9 & 15.9 & \textbf{18.9} & 29.9 & 11.3 & 11.9 & 79.7 & 76.2 & \textbf{54.7} & 10.3 & 79.7 & 29.3 & 17.2 & 14.1 &	37.4 \\ \hline
GTA-only DA & CyCADA~\cite{hoffman2018cycada} & 53.3 & 15.7 & 64.0 & 5.1 & 14.9 & 28.9 & 24.3 & 13.0 & 63.2 & 81.4 & 46.3 & 10.8 & 75.5 & 31.6 & 22.2 & 5.1 & 34.7 \\
\hline
SYNTHIA-only DA & CyCADA~\cite{hoffman2018cycada} & 22.0 & 12.5 & 46.7 & 0.2 & 0.0 & 25.0 & 8.4 & 12.4 & 68.8 & 85.2 & 34.8 & 11.5 & 60.6 & 23.7 & 19.1 & 12.3 & 27.7 \\
\hline
Source-combined DA & CyCADA~\cite{hoffman2018cycada} & 64.9 & 33.6 & 73.3 & 15.8 & 15.3 & 29.2 & 15.9 & 21.4 & 79.3 & 79.0 & 52.0 & 12.7 & 49.7 & 14.0 & 17.5 & 22.5 &	37.2 \\
\hline
& MDAN~\cite{zhao2018adversarial} & 57.6 & 31.2 & 53.5 & 6.5 & 0.6 & 20.3 & 0.0 & 0.0 & 73.0 & 61.7 & 40.9 & 9.8 & 60.4 & 29.2 & 10.3 & 15.6 &	29.4 \\
\multirow{-1}{*}{Multi-source DA}&\textbf{MADAN~(Ours)} & 74.5 & 32.4 & 71.3 & 16.5 & 16.3 & \textbf{30.6} & 15.1 & \textbf{25.1} & \textbf{80.6} & 78.7 & 52.2 & 12.4 & 70.5 & 34.0 & 18.4 & 19.4 &	40.4  \\
&\textbf{MADAN+~(Ours)} &  \textbf{87.8} & \textbf{44.2} & \textbf{78.6} & \textbf{22.4} & 6.8 & 29.1 & 11.5 & 5.3 & 79.6 & 74.6 & 53.6 & \textbf{14.6} & \textbf{83.0} & \textbf{43.4} & 19.1 & \textbf{30.2} & \textbf{42.7} \\
\hline
Oracle-Train on Target & DeepLabV2~\cite{chen2017deeplab} & 93.3 & 59.6 & 82.4 & 28.7 & 45.8 & 40.3 & 42.8 & 43.9 & 84.5 & 94.3 & 60.4 & 24.3 & 87.5 & 74.2 & 45.2 & 51.8 &	59.9  \\
\bottomrule
\end{tabular}
}
\label{tab:comparison_with_SOTA_BDD_Deeplab}
\end{table*}

In Digits-five, Office-31 and Office+Caltech-10 experiments, we use AlexNet \cite{krizhevsky2012imagenet} as our backbone. In Office-Home experiments, we adopt ResNet-50~\cite{he2016deep} as our backbone. In the training stage, we use an Adam optimizer with a batch size of 32 and a learning rate of 1e-3 and 1e-4 respectively for the classification model and feature-level alignment.

%In Office-Home experiments, to compare with the state-of-the-art methods such as~\cite{long2017deep}, we adopt ResNet-50 as our backbone. Other optimization settings are the same as above.

In segmentation adaptation experiments, we choose to use FCN~\cite{long2015fully} as our semantic segmentation network, and, as the VGG family of networks is commonly used in reporting DA results, we use VGG-16~\cite{vgg} as the FCN backbone. The weights of the feature extraction layers in the networks are initialized from models trained on ImageNet~\cite{deng2009imagenet}. The network is implemented in PyTorch and trained with Adam optimizer~\cite{adam} using a batch size of 8 with initial learning rate 1e-4. We keep the image size the same before and after image translation, and crop the adapted images to $400\times400$ during the segmentation model training with 40 epochs. We take the 16 intersection classes of GTA and SYNTHIA, compatible with Cityscapes and BDDS, for all mIoU evaluations. To better illustrate the effectiveness of our proposed model, we also employ DeepLabV2~\cite{chen201 7deeplab} with ResNet-101~\cite{he2016deep} pretrained on ImageNet~\cite{deng2009imagenet} as the semantic segmentation model.

%To better illustrate the effectiveness of our proposed model, we also employ MADAN on DeepLabV2~\cite{chen2017deeplab} model with ResNet-101~\cite{he2016deep} pretrained on ImageNet~\cite{deng2009imagenet} on Dynamic Adversarial Image Generation (DAIG), feature-level Alignment (FLA) and Category level Alignment (CLA). During DAIG, we use Deeplab-V2 model to ensure semantic consistency between source and intermediate domains. During FLA and CLA, we use Resnet-101\cite{he2016deep} to extract features for discriminators and we adopt classifiers in Deeplab-V2 \cite{chen2017deeplab} to perform segmentation task on the generated images.

For digit recognition and object classification, one domain is selected as the target domain and the rest are considered as source domains. For semantic segmentation, we choose synthetic GTA and SYNTHIA as source domains and real Cityscapes and BDDS as target domains.

\subsection{Comparison with State-of-the-art}

Table~\ref{tab:DigitsFive}, Table~\ref{tab:Office31}, Table~\ref{tab:Office-Caltech}, and Table~\ref{tab:Office-Home} show the performance comparisons between the proposed MADAN model and the other baselines, including source-only, single-source DA, source-combined DA, and multi-source DA, on Digits-five, Office-31, Office+Caltech-10, and Office-Home datasets, respectively. The simulation-to-real semantic segmentation adaptation from synthetic GTA and SYNTHIA to real Cityscapes and BDDS are shown in Table~\ref{tab:comparison_with_SOTA_CS_FCN} and Table~\ref{tab:comparison_with_SOTA_BDD_FCN} for FCN-VGG16 backbone, and Table~\ref{tab:comparison_with_SOTA_CS_Deeplab} and Table~\ref{tab:comparison_with_SOTA_BDD_Deeplab} for DeepLabV2-ResNet101 backbone, respectively.
From the results, we have the following similar observations among different adaptation tasks:

\begin{figure*}[!t]
\centering
\includegraphics[width=1.0\linewidth]{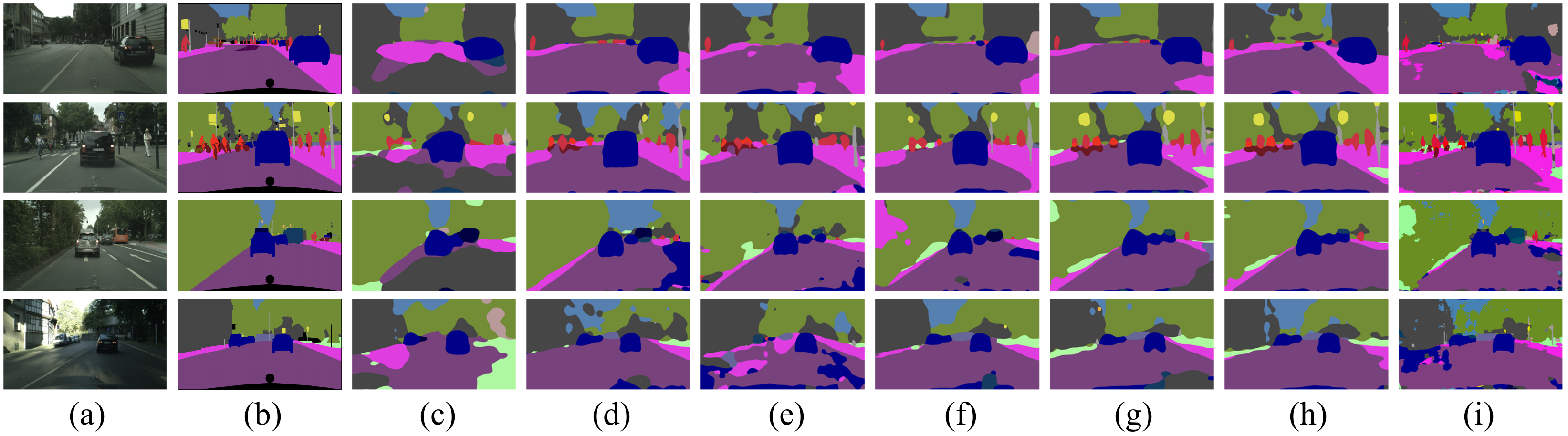}
\caption{Qualitative semantic segmentation result from GTA and SYNTHIA to Cityscapes. From left to right are: (a) original image, (b) ground truth annotation, (c) source only from GTA, (d) CycleGANs on GTA and SYNTHIA, (e) +CCD+DSC, (f) +SAD+DSC, (g) +CCD+SAD+DSC, (h) +CCD+SAD+DSC+FLA~(MADAN), and (i) +CCD+SAD+DSC+FLA+CLA+CAG~(MADAN+).}
\label{fig:SegVisual}
\end{figure*}

\begin{table*}[!t]
\caption{Comparison between the proposed dynamic semantic consistency (DSC) loss in MADAN and the original SC loss in~\cite{hoffman2018cycada} on Cityscapes  using FCN-VGG16 backbone. The better mIoU for each pair is emphasized in bold.}
\resizebox{\textwidth}{!}{%
\begin{tabular}{c|l|cccccccccccccccc|c}
\toprule
\multirow{1}{*}[1.8em]{Source}&\multirow{1}{*}[1.8em]{Method}  & \rot{road} & \rot{sidewalk} & \rot{building} & \rot{wall} & \rot{fence} & \rot{pole} & \rot{t-light} & \rot{t-sign} & \rot{vegettion} & \rot{sky}   & \rot{person} & \rot{rider} & \rot{car} & \rot{bus} & \rot{m-bike} & \rot{bicycle} & \rot{mIoU} \\ \hline
&CycleGAN+SC & 85.6 & 30.7 & 74.7 & 14.4 & 13.0 & 17.6 & 13.7 & 5.8 & 74.6 & 69.9 & 38.2 & 3.5 & 72.3 & 5.0 & 3.6 & 0.0 & 32.7 \\

\cellcolor{white}&CycleGAN+DSC & 76.6 & 26.0 & 76.3 & 17.3 & 18.8 & 13.6 & 13.2 & 17.9 & 78.8 & 63.9 & 47.4 & 14.8 & 72.2 & 24.1 & 19.8 & 10.8 & \textbf{38.1} \\ \hhline{~|-|-|-|-|-|-|-|-|-|-|-|-|-|-|-|-|-|-}
&CyCADA w/ SC & 85.2 & 37.2 & 76.5 & 21.8 & 15.0 & 23.8 & 21.5 & 22.9 & 80.5 & 60.7 & 50.5 & 9.0 & 76.9 & 28.2 & 9.8 & 0.0 & 38.7 \\

\cellcolor{white}\multirow{-4}{*}{GTA}&CyCADA w/ DSC & 84.1 & 27.3 & 78.3 & 21.6 & 18.0 & 13.8 & 14.1 & 16.7 & 78.1 & 66.9 & 47.8 & 15.4 & 78.7 & 23.4 & 22.3 & 14.4 & \textbf{40.0} \\\hline
&CycleGAN+SC & 64.0 & 29.4 & 61.7 & 0.3 & 0.1 & 15.3 & 3.4 & 5.0 & 63.4 & 68.4 & 39.4 & 11.5 & 46.6 & 10.4 & 2.0 & 16.4 & 27.3 \\

\cellcolor{white}&CycleGAN + DSC & 68.4 & 29.0 & 65.2 & 0.6 & 0.0 & 15.0 & 0.1 & 4.0 & 75.1 & 70.6 & 45.0 & 11.0 & 54.9 & 18.2 & 3.9 & 26.7 & \textbf{30.5} \\ \hhline{~|-|-|-|-|-|-|-|-|-|-|-|-|-|-|-|-|-|-}
&CyCADA w/ SC & 66.2 & 29.6 & 65.3 & 0.5 & 0.2 & 15.1 & 4.5 & 6.9 & 67.1 & 68.2 & 42.8 & 14.1 & 51.2 & 12.6 & 2.4 & 20.7 & 29.2 \\

\cellcolor{white}\multirow{-4}{*}{SYNTHIA}&CyCADA w/ DSC & 69.8 & 27.2 & 68.5 & 5.8 & 0.0 & 11.6 & 0.0 & 2.8 & 75.7 & 58.3 & 44.3 & 10.5 & 68.1 & 22.1 & 11.8 & 32.7 & \textbf{31.8} \\
\bottomrule
\end{tabular}
}
\label{tab:ablation_1}
\end{table*}

\begin{table*}[!t]
\caption{Comparison between the proposed dynamic semantic consistency (DSC) loss in MADAN and the original SC loss in~\cite{hoffman2018cycada} on BDDS  using FCN-VGG16 backbone. The better mIoU for each pair is emphasized in bold.}
\resizebox{\textwidth}{!}{%
\begin{tabular}{c|l|cccccccccccccccc|c}
\toprule
\multirow{1}{*}[1.8em]{Source}&\multirow{1}{*}[1.8em]{Method}  & \rot{road} & \rot{sidewalk} & \rot{building} & \rot{wall} & \rot{fence} & \rot{pole} & \rot{t-light} & \rot{t-sign} & \rot{vegettion} & \rot{sky}   & \rot{person} & \rot{rider} & \rot{car} & \rot{bus} & \rot{m-bike} & \rot{bicycle} & \rot{mIoU} \\ \hline
&CycleGAN+SC & 62.1 & 20.9 & 59.2 & 6.0 & 23.5 & 12.8 & 9.2 & 22.4 & 65.9 & 78.4 & 34.7 & 11.4 & 64.4 & 14.2 & 10.9 & 1.9 & 31.1 \\

\cellcolor{white}&CycleGAN+DSC & 74.4 & 23.7 & 65.0 & 8.6 & 17.2 & 10.7 & 14.2 & 19.7 & 59.0 & 82.8 & 36.3 & 19.6 & 69.7 & 4.3 & 17.6 & 4.2 & \textbf{32.9} \\ \hhline{~|-|-|-|-|-|-|-|-|-|-|-|-|-|-|-|-|-|-}
&CyCADA w/ SC & 68.8 & 23.7 & 67.0 & 7.5 & 16.2 & 9.4 & 11.3 & 22.2 & 60.5 & 82.1 & 36.1 & 20.6 & 63.2 & 15.2 & 16.6 & 3.4  & 32.0 \\

\cellcolor{white}\multirow{-4}{*}{GTA}&CyCADA w/ DSC & 70.5 & 32.4 & 68.2 & 10.5 & 17.3 & 18.4 & 16.6 & 21.8 & 65.6 & 82.2 & 38.1 & 16.1 & 73.3 & 20.8 & 12.6 & 3.7 & \textbf{35.5} \\\hline
&CycleGAN+SC & 50.6 & 13.6 & 50.5 & 0.2 & 0.0 & 7.9 & 0.0 & 0.0 & 63.8 & 58.3 & 21.6 & 7.8 & 50.2 & 1.8 & 2.2 & 19.9 & 21.8 \\

\cellcolor{white}&CycleGAN + DSC & 57.3 & 13.4 & 56.1 & 2.7 & 14.1 & 9.8 & 7.7 & 17.1 & 65.5 & 53.1 & 11.4 & 1.4 & 51.4 & 13.9 & 3.9 & 8.7 & \textbf{22.5} \\ \hhline{~|-|-|-|-|-|-|-|-|-|-|-|-|-|-|-|-|-|-}
&CyCADA w/ SC & 49.5 & 11.1 & 46.6 & 0.7 & 0.0 & 10.0 & 0.4 & 7.0 & 61.0 & 74.6 & 17.5 & 7.2 & 50.9 & 5.8 & 13.1 & 4.3 & 23.4 \\

\cellcolor{white}\multirow{-4}{*}{SYNTHIA}&CyCADA w/ DSC & 55.0 & 13.8 & 45.2 & 0.1 & 0.0 & 13.2 & 0.5 & 10.6 & 63.3 & 67.4 & 22.0 & 6.9 & 52.5 & 10.5 & 10.4 & 13.3 & \textbf{24.0} \\
\bottomrule
\end{tabular}
}
\label{tab:ablation_1_BDD}
\end{table*}

\textbf{(1)} The source-only method that directly transfers the task models trained on the source domains to the target domain obtains the worst performance in most adaptation settings. This is obvious, because the joint probability distributions of observed images and labels are significantly different among the sources and the target, due to the presence of domain shift. Without domain adaptation, the direct transfer cannot well handle this domain gap.

%\textbf{(1)} The source-only method, \textit{i.e.} directly transferring the models trained on the source domains to the target domain, performs the worst in most adaptation settings. Due to the presence of \emph{domain shift} or \emph{dataset bias}, the joint probability distributions of observed images and class labels greatly differ between the source and target domains. This results in the model's low transferability from the source domains to the target domain. Simply combining different source domains performs better than each single source, which indicates the superiority of multiple sources over single source despite the domain shift among different sources.

\textbf{(2)} Comparing source-only with corresponding single-best DA and source-combined DA for digit recognition and object classification, and comparing source-only with single-source DA for semantic segmentation, it is clear that almost all adaptation methods perform better than source-only, which demonstrates the effectiveness of domain adaptation. For example, in Table~\ref{tab:Office31}, the average accuracy of source-only combined method is 80.2\%, while the accuracy of source-combined ADDA is 83.7\%.

\begin{table*}[!t]
\caption{Ablation study on different components in MADAN+ on Cityscapes using FCN-VGG16 backbone. Baseline denotes using pixel-level alignment with cycle-consistency, +SAD denotes using the sub-domain aggregation discriminator, +CCD denotes using the cross-domain cycle discriminator, +DSC denotes using the dynamic semantic consistency loss, +FLA denotes using feature-level alignment, +CAG denotes using context-aware generation.}
\resizebox{\textwidth}{!}{%
\begin{tabular}{c|cccccccccccccccc|c}
\toprule
\multirow{1}{*}[1.8em]{Method}  & \rot{road} & \rot{sidewalk} & \rot{building} & \rot{wall} & \rot{fence} & \rot{pole} & \rot{t-light} & \rot{t-sign} & \rot{vegettion} & \rot{sky}   & \rot{person} & \rot{rider} & \rot{car} & \rot{bus} & \rot{m-bike} & \rot{bicycle} & \rot{mIoU} \\ \hline
Baseline & 74.9 & 27.6 & 67.5 & 9.1 & 10.0 & 12.8 & 1.4 & 13.6 & 63.0 & 47.1 & 41.7 & 13.5 & 60.8 & 22.4 & 6.0 & 8.1 & 30.0 \\

+SAD & 79.7 & 33.2 & 75.9 & 11.8 & 3.6 & 15.9 & 8.6 & 15.0 & 74.7 & 78.9 & 44.2 & 17.1 & 68.2 & 24.9 & 16.7 & 14.0 & 36.4 \\
+CCD & 82.1 & 36.3 & 69.8 & 9.5 & 4.9 & 11.8 & 12.5 & 15.3 & 61.3 & 54.1 & 49.7 & 10.0 & 70.7 & 9.7 & 19.7 & 12.4 & 33.1 \\

+SAD+CCD & 82.7 & 35.3 & 76.5 & 15.4 & \textbf{19.4} & 14.1 & 7.2 & 13.9 & 75.3 & 74.2 & 50.9 & 19.0 & 66.5 & 26.6 & 16.3 & 6.7 & 37.5 \\
+SAD+DSC & 83.1 & 36.6 & 78.0 & 23.3 & 12.6 & 11.8 & 3.5 & 11.3 & 75.5 & 74.8 & 42.2 & 17.9 & 72.2 & 27.2 & 13.8 & 10.0 & 37.1 \\

+CCD+DSC & 86.8 & 36.9 & 78.6 & 16.2 & 8.1 & 17.7 & 8.9 & 13.7 & 75.0 & 74.8 & 42.2 & 18.2 & 74.6 & 22.5 & \textbf{22.9} & 12.7 & 38.1 \\

+SAD+CCD+DSC & 84.2 & 35.1 & 78.7 & 17.1 & 18.7 & 15.4 & 15.7 & \textbf{24.1} & 77.9 & 72.0 & 49.2 & 17.1 & 75.2 & 24.1 & 18.9 & 19.2 & 40.2 \\

SAD+CCD+DSC+FLA & 86.2 & 37.7 & 79.1 & 20.1 & 17.8 & 15.5 & 14.5 & 21.4 & 78.5 & 73.4 & 49.7 & 16.8 & 77.8 & 28.3 & 17.7 & \textbf{27.5} & 41.4 \\

+SAD+CCD+DSC+FLA+CLA & 87.7 & \textbf{45.2} & \textbf{80.2} & \textbf{24.0} & 12.4 & 16.0 & 13.4 & 14.8 & \textbf{79.8} & 76.7 & 49.7 & 20.8 & 79.9 & 24.9 & 19.5 & 20.6 & 41.6 \\

+SAD+CCD+DSC+FLA+CLA+CAG & \textbf{87.9} & 41.0 & 76.4 & 21.4 & 1.3 & \textbf{28.4} & \textbf{20.3} & 22.3 & 77.3 & \textbf{80.0} & \textbf{54.9} & \textbf{21.5} & \textbf{80.1} & \textbf{29.7} & 15.1 & 26.5 & \textbf{42.8} \\
\bottomrule
\end{tabular}
}
\label{tab:ablation_2}
\end{table*}

\begin{table*}[!t]
\caption{Ablation study on different components in MADAN+ on BDDS using FCN-VGG16 backbone.}
\resizebox{\textwidth}{!}{%
\begin{tabular}{c|cccccccccccccccc|c}
\toprule
\multirow{1}{*}[1.8em]{Method}  & \rot{road} & \rot{sidewalk} & \rot{building} & \rot{wall} & \rot{fence} & \rot{pole} & \rot{t-light} & \rot{t-sign} & \rot{vegettion} & \rot{sky}   & \rot{person} & \rot{rider} & \rot{car} & \rot{bus} & \rot{m-bike} & \rot{bicycle} & \rot{mIoU} \\ \hline
Baseline & 31.3 & 17.4 & 55.4 & 2.6 & 12.9 & 12.4 & 6.5 & 18.0 & 63.2 & 79.9 & 21.2 & 5.6 & 44.1 & 14.2 & 6.1 & 11.7 & 24.6 \\

+SAD & 58.9 & 18.7 & 61.8 & 6.4 & 10.7 & 17.1 & 20.3 & 17.0 & 67.3 & 83.7 & 21.1 & 6.7 & 66.6 & 22.7 & 4.5 & 14.9 & 31.2 \\
+CCD & 52.7 & 13.6 & 63.0 & 6.6 & 11.2 & 17.8 & 21.5 & 18.9 & 67.4 & \textbf{84.0} & 9.2 & 2.2 & 63.0 & 21.6 & 2.0 & 14.0 & 29.3 \\

+SAD+CCD & 61.6 & 20.2 & 61.7 & 7.2 & 12.1 & 18.5 & 19.8 & 16.7 & 64.2 & 83.2 & 25.9 & 7.3 & 66.8 & 22.2 & 5.3 & 14.9 & 31.8 \\
+SAD+DSC & 60.2 & 29.5 & 66.6 & 16.9 & 10.0 & 16.6 & 10.9 & 16.4 & 78.8 & 75.1 & 47.5 & 17.3 & 48.0 & \textbf{24.0} & 13.2 & 17.3 & 34.3 \\

+CCD+DSC &61.5 & 27.6 & 72.1 & 6.5 & 12.8 & 15.7 & 10.8 & 18.1 & 78.3 & 73.8 & 44.9 & 16.3 & 41.5 & 21.1 & 21.8 & 15.9 & 33.7 \\
+SAD+CCD+DSC & 64.6 & \textbf{38.0} & 75.8 & 17.8 & 13.0 & 9.8 & 5.9 & 4.6 & 74.8 & 76.9 & 41.8 & \textbf{24.0} & 69.0 & 20.4 & 23.7 & 11.3 & 35.3 \\

+SAD+CCD+DSC+FLA & 69.1 & 36.3 & 77.9 & 21.5 & 17.4 & 13.8 & 4.1 & 16.2 & 76.5 & 76.2 & 42.2 & 16.4 & 56.3 & 22.4 & \textbf{24.5} & 13.5 & 36.3 \\

+SAD+CCD+DSC+FLA+CLA & 75.1 & 30.5 & 70.8 & 10.3 & 11.5 & 27.8 & 10.6 & 15.9 & 80.6 & 80.9 & \textbf{51.0} & 12.2 & 67.2 & 21.3 & 17.2 & \textbf{22.4} & 37.8 \\
+SAD+CCD+DSC+FLA+CLA+CAG &  \textbf{75.2} & 29.8 & \textbf{83.3} & \textbf{27.2} & \textbf{20.7} & \textbf{37.8} & \textbf{23.2} & \textbf{20.6} & \textbf{81.1} & 83.5 & 50.1 & 9.8 & \textbf{80.2} & 13.2 & 11.6 & 18.1 & \textbf{41.6} \\
\bottomrule
\end{tabular}
}
\label{tab:ablation_2_BDD}
\end{table*}

\textbf{(3)} Generally, multi-source DA outperforms other adaptation standards by exploring the complementarity of different sources. This is more obvious when comparing the DA methods that employ similar architectures, such as our MADAN vs. CyCADA~\cite{hoffman2018cycada}, MDDA~\cite{zhao2020multi} vs. ADDA~\cite{tzeng2017adversarial}, and MDAN~\cite{zhao2018adversarial} vs. DANN~\cite{ganin2016domain}. Besides the domain gap between the sources and the target, multi-source DA also tries to bridge the domain gap across different sources. This demonstrates the necessity and superiority of multi-source DA over single-source DA.

\textbf{(4)} MADAN achieves the best average results among all adaptation methods, benefiting from the joint consideration of pixel-level and feature-level alignments, cycle-consistency, dynamic semantic consistency, domain aggregation, and multiple sources.
MADAN also significantly outperforms source-combined DA, in which domain shift also exists among different sources. By bridging this gap, multi-source DA can boost the adaptation performance.
On the one hand, compared to single-source DA like CyCADA~\cite{hoffman2018cycada}, MADAN utilizes more useful information from multiple sources. On the other hand, other multi-source DA methods~\cite{xu2018deep,zhao2018adversarial,peng2019moment,zhao2020multi} only consider feature-level alignment, which is obviously insufficient especially for fine-grained tasks, \textit{e.g.} semantic segmentation, a pixel-wise prediction task. In addition, we consider pixel-level alignment with a dynamic semantic consistency loss and further aggregate different adapted domains.

\textbf{(5)} Take segmentation segmentation for example, the oracle method that is trained on the target domain performs significantly better than the others. However, to train this model, the ground truth labels from the target domain are required, which are actually unavailable in UDA settings. We can deem this performance as a upper bound of UDA. Obviously, there is still a large performance gap between all adaptation algorithms and the oracle method, requiring further efforts on DA.

%\textbf{(4)} The oracle method, \textit{i.e.} testing on the target domain using the model trained on the same domain, achieves the best performance. However, this model is trained using the ground truth segmentation labels from the target domain, which are unavailable in unsupervised domain adaptation. Obviously, there is still a large performance gap between all adaptation methods and the oracle method, requiring further efforts on DA.  %Specifically,

% \begin{table*}[]
% \caption{Class-wise Performance comparison from GTA to Cityscapes~(\textbf{19 classes)}. }
% \vspace{1mm}

% \resizebox{\textwidth}{!}{%
% \begin{tabular}{l|ccccccccccccccccccc|c}
% \toprule[1.5pt]
% \multirow{1}{*}[1.8em]{Method} & \rot{road} & \rot{sidewalk} & \rot{building} & \rot{wall} & \rot{fence} & \rot{pole} & \rot{t-light} & \rot{t-sign} & \rot{vegettion} & \rot{terrain} & \rot{sky}   & \rot{person} & \rot{rider} & \rot{car}  & \rot{truck} & \rot{bus}  & \rot{train} & \rot{m-bike} & \rot{bicycle} & \rot{mIoU} \\ \hline

% NonAdapt~\cite{zhang2017curriculum} & & & & & & & & & & & & & & & & & & & & \\
% \hline
% \end{tabular}
% }
% \vspace{-2mm}
% \label{tab:adaptation_gta}
% \end{table*}

There are also some task-specific observations:

\textbf{(1)} Simply combining different source domains into one source and performing source-only or single-source DA does not guarantee better performance than corresponding single-best method. For example, for the source-only standard, the single-best method outperforms the combined method on Digits-five, Office-31, Office+Caltech-10 datasets, while the combined method performs better on Office-Home, Cityscapes, and BDDS datasets. For the single-source DA, we usually have opposite observations. For example, in Table~\ref{tab:comparison_with_SOTA_CS_FCN}, the mIoUs of CyCADA from GTA to Cityscapes and from SYNTHIA to Cityscapes are 38.7\% and 29.2\%, while the mIoU of source-combined DA is 37.3\%. Currently, there is no accurate explanation on this observation. On the one hand, combining multiple sources into one source results in more training data, which can intuitively boost the performance. On the other hand, the data from different sources are collected from different distributions, which may interfere with each other. Therefore, the comparison between the single-best method and the combined method depends on which aspect is stronger.

(2) For semantic segmentation adaptation, MADAN+ outperforms MADAN with a remarkable margin. For example, the average performance gains of MADAN+ over MADAN using DeepLabV2 backbone are 3.1\% and 2.3\% on Cityscapges and BDDS, respectively. Further, MADAN+ achieves the best cwIoU scores of 6 to 9 out of 16 categories. These results demonstrate the superiority of MDAN+ over MADAN for pixel-wise segmentation adaptation with the help of category-level alignment and context-aware generation.

\noindent\textbf{Segmentation Visualization.} The qualitative semantic segmentation results are shown in Figure~\ref{fig:SegVisual}. We can clearly see that after adaptation by the proposed method, the visual segmentation results are improved notably, which look more similar to the ground truth (b). Take the second row for example, the contours of pedestrians and cyclists by MADAN+ (i) are more clear than those by the methods of source only (c) and CycleGAN (d).

\begin{figure*}[!t]
\begin{center}
\centering \includegraphics[width=0.95\linewidth]{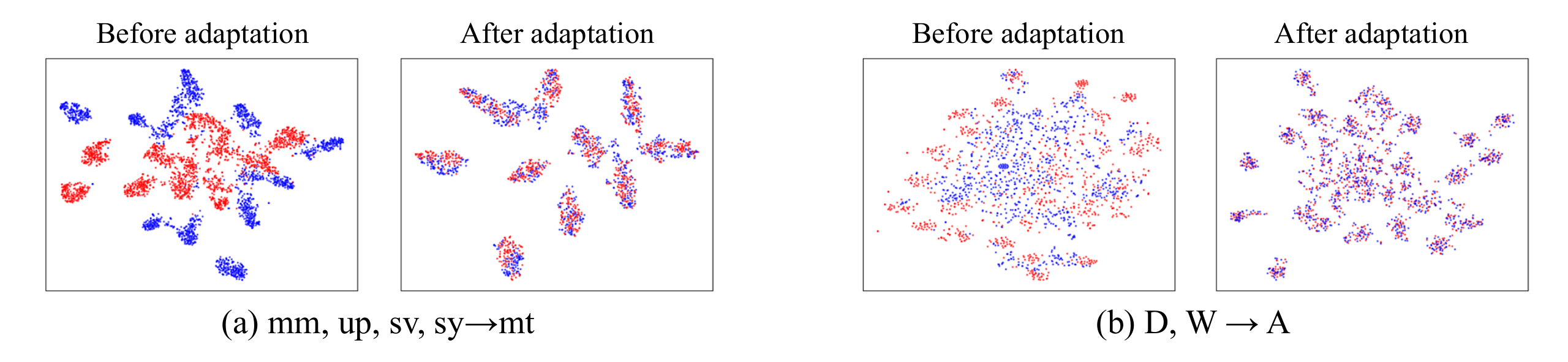}
\caption{The t-SNE~\cite{maaten2008visualizing} visualization of the learned features for task (a) Digits-five: mm, up, sv, sy$\rightarrow$mt and (b) Office-31: D, W$\rightarrow$A. In each pair, the features are extracted using the last layer of source domain encoder from the samples of source and target domain in the first image, and the target domain features are extracted using the the last layer of adapted encoder in the second one.}
\label{fig:tSNE}
\end{center}
\end{figure*}

\begin{figure*}[!t]
\centering
\includegraphics[width=1.0\linewidth]{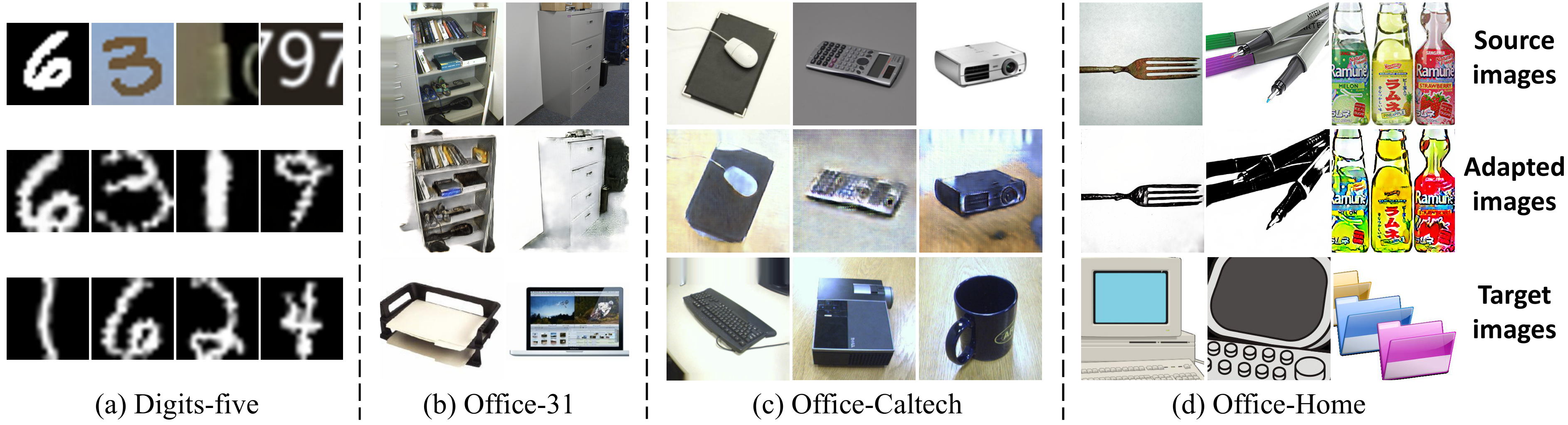}
\caption{Visualization of image translation for classification adaptation. From left to right are: (a) Digits-five: mt, mm, sv, sy $\rightarrow$ up,  (b) Office-31: W, D $\rightarrow$ A, (c), Office+Caltech-10: D, C, A $\rightarrow$ W (d) Office-Home: Ar, Rw, Pr $\rightarrow$ Cl. Red: source, blue: target.}
\label{fig:AdaptedImages_Classification}
\end{figure*}

\begin{figure*}[!t]
\centering
\includegraphics[width=1.0\linewidth]{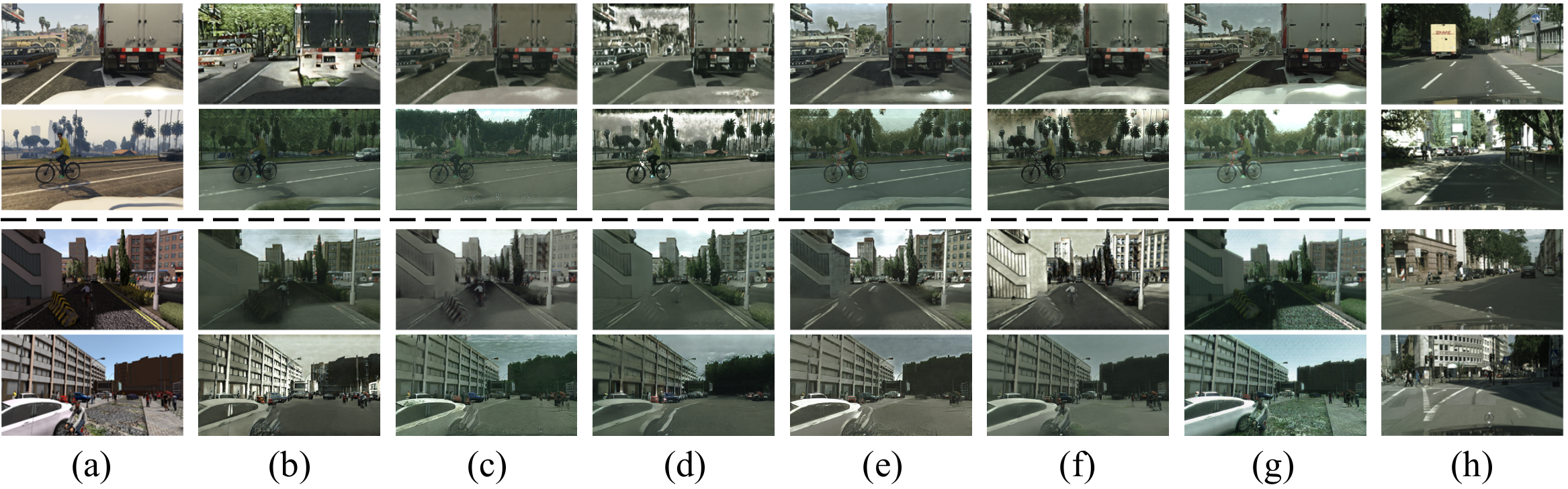}
\caption{Visualization of image translation for segmentation adaptation from GTA and SYNTHIA to Cityscapes. From left to right are: (a) original source image, (b) CycleGAN, (c) CycleGAN+DSC, (d) CycleGAN+CCD+DSC, (e) CycleGAN+SAD+DSC, (f) CycleGAN+CCD+SAD+DSC, (g) CycleGAN+CCD+SAD+CAG, and (h) target Cityscapes image. The top two rows and bottom rows are GTA $\rightarrow$ Cityscapes and SYNTHIA $\rightarrow$  Cityscapes, respectively.}
\label{fig:AdaptedImages_Segmentation}
\end{figure*}

\subsection{Ablation Study}
To demonstrate the effectiveness of different components in the proposed MADAN and MADAN+ models, we conduct ablation studies on the segmentation adaptation tasks.

First, we compare the proposed dynamic semantic consistency (DSC) loss with the original semantic consistency (SC) loss~\cite{hoffman2018cycada} using the DA methods of CycleGAN~\cite{zhu2017unpaired} and CyCADA~\cite{hoffman2018cycada}. The results on Cityscapes and BDDS are shown in Table~\ref{tab:ablation_1} and Table~\ref{tab:ablation_1_BDD}, respectively. We can see that for all adaptation settings, DSC achieves better mIoU results than SC. For example, the mIoU improvements of DSC over SC in CycleGAN and CyCADA from GTA to Cityscapes are 5.4\% and 1.3\%, respectively, while the corresponding improvements are 3.2\%  and 2.6\% from SYNTHIA to Cityscapes. These results demonstrate the effectiveness of our proposed DSC loss.

%After demonstrating its value, we employ the DSC loss in subsequent experiments.

Second, we incrementally evaluate the influence of different components in MADAN+.
The results on Cityscapes and BDDS using FCN-VGG16 backbone are shown in Table~\ref{tab:ablation_2} and Table~\ref{tab:ablation_2_BDD}, respectively. We have several observations. (1) Both domain aggregation methods, \textit{i.e.} SAD and CCD, obtain larger mIoU scores than baseline with SAD performing better. The performance gains are obtained by making different adapted domains more closely aggregated. (2) Adding the DSC loss could further improve the segmentation performance, again demonstrating the effectiveness of DSC. (3) feature-level alignment is also helpful with 1.2\% and 1.0\% improvements on Cityscapes and BDDS, respectively, obviously contributing to the adaptation task. (4) Category-level alignment (CLA) is complementary to the feature-level alignment (FLA). While FLA aims to align the target and source features globally, CLA makes the features in local regions indistinguishable. (5) context-aware generation (CAG) significantly contributes to the adaptation task. (6) The modules are orthogonal to each other to some extent, since adding each one of them does not introduce performance degradation. (7) As compared to MADAN, MADAN+ achieves better results with 1.4\% and 5.3\% performance gains on Cityscapes and BDDS, respectively. Moreover, by adding CLA and CAG, the cwIoU of most categories are increased. These results demonstrate the superiority of MADAN+ over MADAN for pixel-wise adaptation.

\begin{figure*}[!t]
\begin{center}
\centering \includegraphics[width=1.0\linewidth]{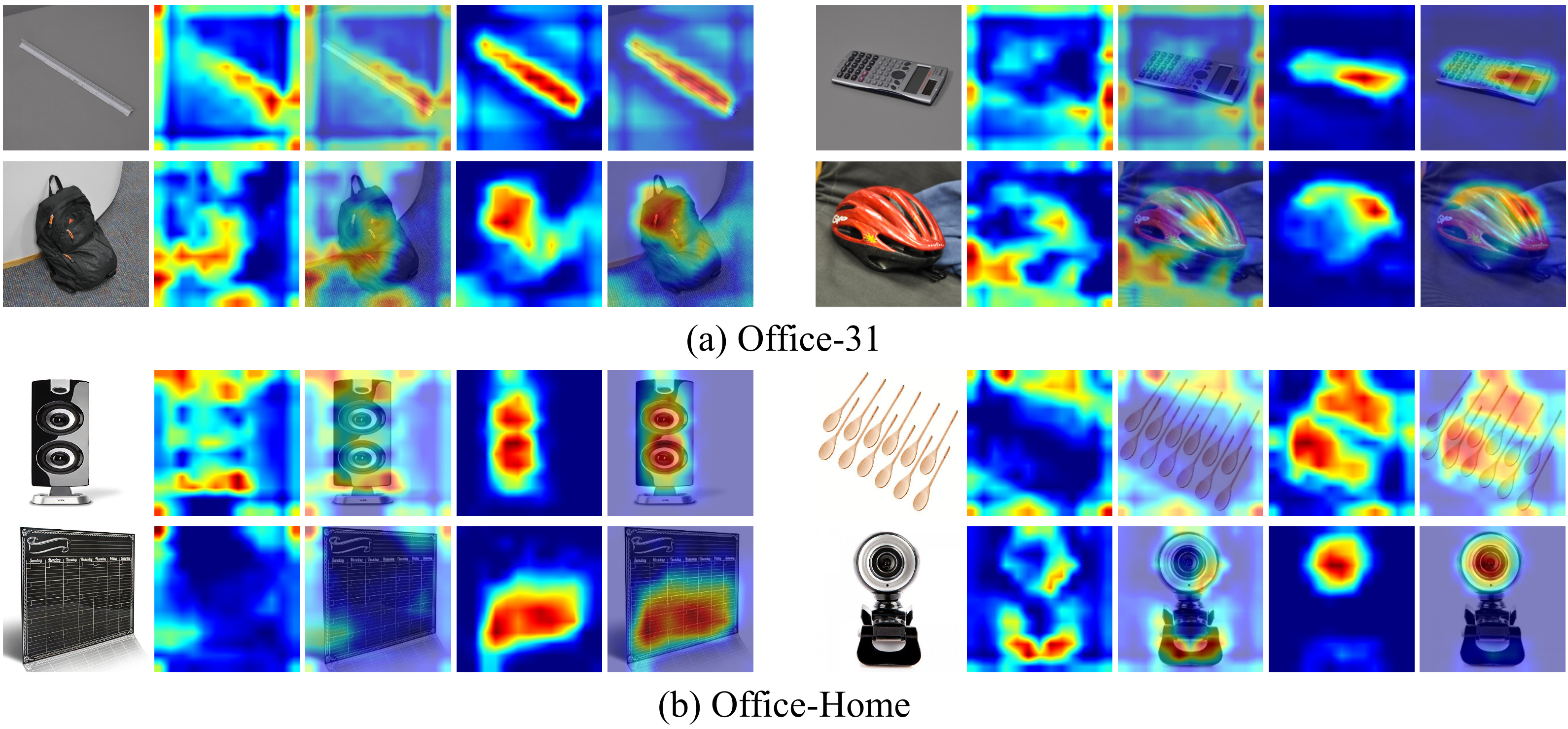}
\caption{Comparison of the attention maps before and after adaptation on (a) Office-31 and (b) Office-Home datasets. For each group, the five columns from left to right are: the original target image, attention map before adaptation, image with attention map before adaptation, attention map after adaptation, and image with attention map after adaptation. Red regions indicate more attention.}
\label{fig:attention}
\end{center}
\end{figure*}

\subsection{Feature Visualization}
To show the feature transferability of the proposed MADAN model, we visualize the features before and after adaptation with t-SNE embedding~\cite{maaten2008visualizing} in two tasks: (a) Digits-five: mm, up, sv, sy$\rightarrow$mt and (b) Office-31: D, W$\rightarrow$A. As illustrated in Figure~\ref{fig:tSNE}, we can observe that after adaptation, the target domain is more indistinguishable from the source domains, which demonstrates that the proposed MADAN model can align the distributions between the source and target domains. Based on the more transferable features after adaptation, the task classifier learned on the source domains can work well on the target domain, leading high task performance on the target domain.

\subsection{Model Interpretability}
We visualize the results of pixel-level alignment (PLA) and attention maps before and after adaptation to demonstrate the interpretability of our model. First, we show the comparison among source image, adapted images, and target images for classification and segmentation adaptation in Figure~\ref{fig:AdaptedImages_Classification} and Figure~\ref{fig:AdaptedImages_Segmentation}, respectively. We can see that the styles of the adapted images by our PLA method are closer to the target than the source to the target. Meanwhile, the semantic information is well preserved. For classification in Figure~\ref{fig:AdaptedImages_Classification}: (a) although styles of the source images are different, the corresponding adapted images are uniformly changed to the handwritten brush style of the target images; (b) the background is removed in the adapted images; (c) a desktop background is added to the adapted images; (d) the adapted images are cartooned to have similar styles to the target images. For segmentation in
Figure~\ref{fig:AdaptedImages_Segmentation},  comparing the columns from (a) to (g) with the column (h) especially (a) vs. (h) and (g) vs. (h), we can observe that with our final FLA method (g), the styles (\textit{e.g.} overall hue and brightness) of the adapted images are much more similar to the target Cityscapes.

Second, we visualize the attention before and after the proposed domain adaptation method using the heat map generated by the Grad-Cam algorithm~\cite{gradcam2017iccv}. The comparison before and after adaptation on Office-31 and Office-Home datasets are illustrated in Figure~\ref{fig:attention}.  It is clear that different regions in the images have different attentions but the attentions generated by our domain adaptation method can focus more on the desirable and discriminative regions. For example, on the Office-31 dataset, for the image in the top right group, the calculator is highlighted with more attention after adaptation, while more attention is focused on a region in the background before adaptation; for the image in the bottom right group, after adaptation more attention is paid to the helmet and the attention diminishes for the complex background with messy objects. On the Office-Home dataset, for the image in the top left group, the attention before adaptation focuses on the background and the edge of the speaker, while the more discriminative and transferable trumpets are emphasized after adaptation; for the image in the bottom right group, only the lens of the Webcam is highlighted after adaptation since it is more transferable than the base of the camera.
These observations intuitively demonstrate that the attended regions by our adaptation model are invariant across different domains and discriminative for the learning task.

% \subsection{Discussions}
% \textbf{Computation cost.} Since the proposed framework deals with a harder problem, \textit{i.e.} multi-source domain adaptation, more modules are used to align different sources, which results in a larger model. In our experiments, MADAN is trained on 4 NVIDIA Tesla P40 GPUs for 40 hours using two source domains which is about twice the training time as on a single source. However, MADAN does not introduce any additional computation during inference, which is the biggest concern in real industrial applications, \textit{e.g.} autonomous driving.

% \textbf{On the poorly performing classes in the segmentation adaptation.} From the results in Table~\ref{tab:comparison_with_SOTA_CS_FCN}, Table~\ref{tab:comparison_with_SOTA_BDD_FCN}, Table~\ref{tab:comparison_with_SOTA_CS_Deeplab}, and Table~\ref{tab:comparison_with_SOTA_BDD_Deeplab}, we can find that MADAN and MADAN+ perform poorly on some classes as other methods, such as \textit{fence} and \textit{pole}. There are two main reasons for the poor performance on certain classes: (1) lack of images containing these classes and (2) structural differences of objects between simulation images and real images (\textit{e.g.} the trees in simulation images are much taller than those in real images). Generating more images for different classes and improving the diversity of objects in the simulation environment are two promising directions for us to explore in future work that may help with these problems.

\section{Conclusion}
\label{sec:Conclusion}
In this paper, we proposed a novel framework, termed Multi-source Adversarial Domain Aggregation Network (MADAN), for multi-source domain adaptation (MDA).
%with multiple sources.
%MADAN is composed of three components, \textit{i.e.} dynamic adversarial image generation, adversarial domain aggregation, and feature-aligned task learning.
For each source domain, based on cycle-consistent GAN at pixel-level alignment, we first generated adapted images with a novel dynamic semantic consistency loss. Further, we proposed a sub-domain aggregation discriminator and cross-domain cycle discriminator to better aggregate different adapted domains. Finally, we trained the task model using the adapted images in the aggregated domain and corresponding labels in the source domains. The experiments showed that MADAN achieves 2.8\%, 3.0\%, 2.2\%, and 4.6\% classification accuracy improvements compared with the existing best MDA methods, respectively on Digits-five, Office-31, Office+Caltech-10, and Office-Home datasets. We also studied MDA for semantic segmentation, which is the first work on adapting pixel-wise prediction task with multiple sources. To better deal with the pixel-wise adaptation, we extended MDAN to MADAN+ with category-level alignment and context-aware generation. For the FCN-VGG16 backbone, MADAN+ achieves 17.0\%, 3.0\%, 5.5\%, and 13.4\% mIoU improvements compared with best source-only, best single-source DA, source-combined DA, and other multi-source DA, respectively on Cityscapes from GTA and SYNTHIA, and 17.0\%, 5.9\%, 7.9\%, 16.6\% on BDDS.

For future studies, we plan to investigate multi-modal DA, such as using both image and LiDAR data, to further boost the adaptation performance. Improving the computational efficiency of MADAN, with techniques such as neural architecture search, is another  direction worth investigating. In addition, we will study how to automatically weigh the relative importance of different sources and the samples in each source to further improve the performance of MADAN.

% use section* for acknowledgment
\ifCLASSOPTIONcompsoc
  % The Computer Society usually uses the plural form
  \section*{Acknowledgments}
\else
  % regular IEEE prefers the singular form
  \section*{Acknowledgment}
\fi
This work is supported by Berkeley DeepDrive and the National Natural Science Foundation of China (No. 61701273).

\ifCLASSOPTIONcaptionsoff
  \newpage
\fi

\bibliographystyle{IEEEtranN}\small
\bibliography{MADAN_Refe}  % sigproc.bib is the name of the Bibliography in this case

% \begin{IEEEbiography}[{\includegraphics[width=1in,height=1.25in,clip,keepaspectratio]{authors/SichengZhao.jpg}}]{Sicheng Zhao}
% received the Ph.D.\ degree from Harbin Institute of Technology, Harbin, China, in 2016. He has been a visiting scholar at National University of Singapore from 2013 to 2014 and a research fellow at Tsinghua University from 2016 to 2017. He is currently a research fellow at University of California Berkeley, USA. His research interests include affective computing, multimedia, and computer vision.
% \end{IEEEbiography}

% \begin{IEEEbiography}[{\includegraphics[width=1in,height=1.25in,clip,keepaspectratio]{authors/KurtKeutzer.jpg}}]{Kurt Keutzer}
% received his Ph.D.\ degree in Computer Science from Indiana University in 1984 and then joined the research division of AT\&T Bell Laboratories. In 1991 he joined Synopsys, Inc.\ where he ultimately became Chief Technical Officer and Senior Vice-President of Research. In 1998, Kurt became Professor of Electrical Engineering and Computer Science at the University of California at Berkeley. Kurt's research group is currently focused on using parallelism to accelerate the training and deployment of Deep Neural Networks for applications in computer vision, speech recognition, multi-media analysis, and computational finance. Kurt has published six books, over 250 refereed articles, and is among the most highly cited authors in Hardware and Design Automation. Kurt is a Fellow of the IEEE.
% \end{IEEEbiography}

% that's all folks
\end{document}